\documentclass[letterpaper, 10 pt, conference]{ieeeconf}  

\IEEEoverridecommandlockouts                              

\usepackage{times}
\usepackage{epsfig}
\usepackage{graphicx}
\usepackage{amsmath}
\usepackage{amssymb}
\usepackage{amsfonts}
\usepackage{multirow}
\usepackage{booktabs}    
\usepackage{chngpage}
\usepackage{xcolor}
\usepackage{soul}
\usepackage{hyperref}
\usepackage[flushleft,para]{threeparttable}

\hypersetup{
    colorlinks=true,
    linkcolor=blue,    
    urlcolor=cyan
}

\newcommand{\real}{\mathbb{R}}

\newcommand{\figref}[1]{Fig.~\ref{#1}}
\newcommand{\tabref}[1]{Table~\ref{#1}}
\newcommand{\secref}[1]{Sec.~\ref{#1}}
\newcommand{\eqnref}[1]{Eqn.~\ref{#1}}

\title{\LARGE \bf
An Adaptive Framework For Learning Unsupervised Depth Completion
}

\author{Alex Wong$^{1}$, Xiaohan Fei$^{2}$, Byung-Woo Hong$^{3}$, and Stefano Soatto$^{1}$
\thanks{This work was supported by ONR N00014-19-1-2229, ARO W911NF-17-1-0304, NRF-2017R1A2B4006023, and NRF-2018R1A4A1059731}
\thanks{$^{1}$Alex Wong and Stefano Soatto are with Department of Computer Science,
        University of California, Los Angeles. Email:
        {\tt\small alexw@cs.ucla.edu}, {\tt\small soatto@cs.sucla.edu}}
\thanks{$^{2}$Xiaohan Fei was with Department of Computer Science,
        University of California, Los Angeles when the work was conducted and is now with Amazon Web Services. Email:
        {\tt\small feixh@cs.ucla.edu}}%
\thanks{$^{3}$Byung-Woo Hong is with the Department of Computer Science,
        Chung-Ang University, Korea. Email:
        {\tt\small hong@cau.ac.kr}}%
}

\begin{document}

\maketitle
\thispagestyle{empty}
\pagestyle{empty}

\begin{abstract}
We present a method to infer a dense depth map from a color image and associated sparse depth measurements. Our main contribution lies in the design of an annealing process for determining co-visibility (occlusions, disocclusions) and the degree of regularization to impose on the model. We show that regularization and co-visibility are related via the fitness (residual) of model to data and both can be unified into a single framework to improve the learning process. Our method is an adaptive weighting scheme that guides optimization by measuring the residual at each pixel location over each training step for (i) estimating a soft visibility mask and (ii) determining the amount of regularization. We demonstrate the effectiveness our method by applying it to several recent unsupervised depth completion methods and improving their performance on public benchmark datasets, without incurring additional trainable parameters or increase in inference time. Code available at: \\ \hyperlink{https://github.com/alexklwong/adaframe-depth-completion}{https://github.com/alexklwong/adaframe-depth-completion}
\end{abstract}

\section{Introduction}
Inferring scene geometry from images supports a variety of tasks, from robotic navigation to image-based rendering. We focus on depth completion, the process of inferring a dense depth map at each instant of time, given an image and sparse depth measurements, which may be obtained from the same image(s) over time via structure-from-motion (SFM), or from a secondary sensor such as a lidar. This is an ill-posed problem, so the solution hinges on the choice of regularization or prior assumptions on the scene. The data fidelity criterion is the usual reprojection error customary in stereo and SFM and subject to visibility phenomena, occlusion and disocclusion. The regularizer imposes generic properties of the scene, for instance piece-wise smoothness and local connectivity. 

There are two distinct phenomena where the data fidelity term (or reprojection error) does not meaningfully constrain the depth map to be inferred: Occlusions, and homogeneous regions. In the latter, there exists a wide range of disparities, one typically chooses the ``simplest'' as defined by the regularizer, for instance the smoothest depth map. In the former, no correct disparity map can fit the data term, since there is no displacement of one image that can match the other. Since no correct disparity exists, one should not penalize the reprojection error in the occluded regions, leaving the depth undefined. Both of these phenomena should be captured, ideally in a \textit{unified fashion}. The main difference is that, whereas in homogeneous regions the data fidelity is already minimized, so the influence of the regularizer is increased automatically, in occluded regions the data fidelity term is uninformative and should be actively ignored so depth information should come from adjacent areas. Our goal is to devise an adaptive unsupervised learning framework that addresses both and fosters this process automatically.

The core of our approach is an adaptive weighting scheme that varies over space (image domain) and time (training steps) and informs (i) the probability of a given pixel being co-visible in two views (for weighting data fidelity) and (ii) the extent in which the prior assumptions (regularization) should be imposed -- driven by the evidence in the data. To account for occlusions and disocclusions, we measure the fitness (residual) of the model to the data at each spatial position over each training time step. The result is a spatially varying soft visibility mask, relevant for spatial tasks such as navigation and manipulation, that \textit{adapts} to the model over training time. The same residual can be used to determine the degree of regularization to impose on each spatial prediction, enabling a second set of adaptive weights. What makes this effective is the fact that, while the regularizers are generic (not informed by large image datasets), the way they are applied is driven by the evidence in the images, which leverages their strength (mostly simplicity) where appropriate, and limits the damage from their simplistic nature where necessary (e.g. across occluding boundaries). Together, the two sets of weights complement each other (i.e. occluded region requires regularization) and are combined into a single framework that can be generically applied to improve both \textit{existing} and \textit{yet-to-be-developed} unsupervised depth completion methods to guide their learning (optimization) to local minima that are more compatible with the data. 

Counter to current trends, our framework requires \textit{no extra trainable parameters}. It is entirely data-driven, leveraging information from the intermediate fitness between model and data as an adaptation signal for both sets of weights. It adaptively weights the data fidelity and regularization terms in the objective function during training and hence incurs \textit{no additional run-time} during inference. Yet, our framework is able to consistently improve the performance of several recent unsupervised depth completion algorithms across public benchmarks, such as KITTI \cite{uhrig2017sparsity} and VOID \cite{wong2020unsupervised}, and achieving new state of the art -- thus, demonstrating its effectiveness. To test the limits of our approach, we also provide a study on  the model performance with lower density of the sparse points. Even with very few (0.05\% density) points, our approach can \textit{still} improve exisiting depth completion methods. 

Our \textbf{contributions} are: (i) an annealed visibility mask that considers the fitness of model to data for determining and discounting occlusions  during training, (ii) the use of residuals from multiple sensor modalities (image and depth) to determine the degree of regularization, and (iii) a unified framework that combines the adaptive weights for discounting occlusions and determining regularization where each set of weights plays a complementary role to the other; (iv) we show that our framework can be generically applied to unsupervised depth completion methods to achieve better performance without incurring additional trainable parameter or run-time complexity during inference.

\section{Related Work}
\noindent \textbf{Supervised Depth Completion.} 
Existing methods regress dense depth from an image and a sparse depth map by minimizing the difference between predictions and ground truth. 
\cite{eldesokey2018propagating} computed confidence from convolutions and propagates it through the layers.
\cite{huang2019hms} performed upsampling followed by convolution to fill the missing values.
\cite{jaritz2018sparse,ma2019self,yang2019dense} used two branches to process image and sparse depth separately. \cite{ma2019self} used early fusion with a ResNet encoder, while \cite{yang2019dense} used late fusion. \cite{jaritz2018sparse} also used late fusion, but with NASNet encoders and jointly learned depth and semantic segmentation. \cite{chen2019learning} proposed a 2D-3D fusion network.
\cite{qu2021bayesian,qu2020depth,van2019sparse} learned confidence maps for guidance and \cite{xu2019depth,zhang2018deep} also used surface normals.
\cite{chodosh2018deep} formulated the problem as compressive sensing and \cite{dimitrievski2018learning} as morphological operators. \cite{merrill2021robust,sartipi2020deep,zuo2021codevio} proposed light-weight networks that can be deployed onto SLAM/VIO systems.

All of these methods are supervised. 
They require ground-truth, often unavailable, or the product of  post-processing and aggregation over a number of consecutive frames~\cite{uhrig2017sparsity}.
Such supervision is not scalable; instead, we learn to predict dense depth by fusing information from the abundant un-annotated images and sparse depth data.

\noindent{\bf Unsupervised (Self-supervised) Depth Completion.}
Unsupervised methods learn depth by minimizing the discrepancy between prediction and sparse depth input, and between the given image and its reconstructions from additional (stereo or temporally adjacent) frames that are available only during training.
Stereo methods \cite{shivakumar2019dfusenet,yang2019dense} predict disparity to reconstruct the given image from its stereo-counterpart and synthesize depth from focal length and baseline. These methods are generally limited to outdoor scenarios.
Monocular methods \cite{ma2019self,wong2021learning,wong2020unsupervised} jointly learn depth and pose by projecting from temporally adjacent frames to a given image. 

As depth completion is an ill-posed problem, regularization is needed. \cite{ma2019self,shivakumar2019dfusenet,wong2020unsupervised} used a generic local smoothness prior that is static with respect to the spatial domain of image and the temporal domain of optimization.
Whereas, \cite{yang2019dense} utilized a \textit{learned prior} (a separate network trained on ground-truth depth) to regularize predictions.  \cite{wong2021learning} learned a topology prior on the sparse points from synthetic data and used it as regularization. We note that supervision from a network trained on a specific domain (e.g. outdoors) will not generalize (e.g. indoors) -- defeating the purpose of unsupervised methods.
Hence, we forgo the use of a learned prior, but instead propose a generic form of regularization that incorporates the local fitness of the current model estimate to data. Unlike conventional regularization, our approach is a locally adaptive, data-driven weighting scheme that varies in space and time and optimization to more desirable local minima. 

\noindent \textbf{Adaptive Weighting Schemes.} 
Many imaging problems are cast into the optimization of an energy function that consists of data fidelity and regularization, where their relative significance is typically determined by a \textit{static} scalar, which often leads to undesirable local minima due to \textit{heteroscedasticity} of residual measuring a discrepancy between model and data.
\cite{galatsanos1992methods} determined the regularization parameter based on noise variance, and \cite{nguyen2001efficient} on the cross-validation criterion.
For depth completion, \cite{ma2019self,shivakumar2019dfusenet,wong2020unsupervised} determines the degree of regularization based on the the image gradient. However, this weighting scheme is still \textit{static} with respect to a given image.
\cite{hong2017adaptive,hong2019adaptive,wong2019bilateral} proposed adaptive regularization in the spatial domain and over the course of optimization based on the local residual. However, their method considers only a single frame. In contrast, we propose an adaptive data-driven algorithm that deals with multiple frames obtained from multiple sensor modalities (image and depth).

Unlike previous works, our method also considers occlusions and disocclusions in the data term.   \cite{ma2019self,shivakumar2019dfusenet,wong2020unsupervised,yang2019dense} uniformly penalized all predictions without accounting for them. Unsupervised monocular depth prediction methods \cite{mahjourian2018unsupervised,wong2020targeted,yang2018every,zhou2017unsupervised} used an extra network to explicitly learn visibility masks by jointly minimizing an unsupervised photometric loss and a penalty for the cardinality of the mask (to avoid degenerate solution of all zeros). We discount unresolved residuals (due to visibility) over the course of optimization \textit{without} incurring an extra network nor training time.

\textbf{Uncertainty in Estimation}. Our work is related to measuring uncertainty for 3D reconstruction. \cite{duggal2019deeppruner,shaked2017improved} proposed to learn uncertainty from groundtruth for stereo. \cite{seki2016patch,spyropoulos2014learning} learned confidences based on deviation from median disparity. \cite{klodt2018supervising} did so in structure-from-motion (SfM) by leveraging existing SfM systems and \cite{mahjourian2018unsupervised,yang2018every,zhou2017unsupervised} in monocular depth prediction. Unlike them, we showed that uncertainty or confidence does not need to be learned, but can be observed given the data. Our work is more in line with classic stereo works \cite{hu2012quantitative} in using the matching cost as a confidence measure, but unlike them, we used it to guide learning.

\section{Motivation}
\label{sec:motivation}
Our goal is to recover a 3D scene from an RGB image $I : \Omega \subset \real^2 \mapsto \real^3_+$ and its associated sparse depth measures $z : \Omega_{z} \subset \Omega \mapsto \real_+$ in an unsupervised learning framework, where depth information is inferred by exploiting additional stereo imagery \cite{shivakumar2019dfusenet,yang2019dense} or temporally adjacent frames \cite{ma2019self,wong2020unsupervised} during the training phase. 
In this work, we assume that temporally adjacent frames, $I_\tau$ for $\tau \in \{ -1, +1 \}$ where $I_{-1}$ denotes the previous frame and $I_{+1}$ the next one with respect to $I$, are available. 
Thus, a training example comprises $(I, I_\tau, z)$. Note: our method can easily be extended to stereo training.
To learn depth, unsupervised depth completion methods minimize a loss function $\mathcal{L}$ that mainly consists of data fidelity $\mathcal{D}$ and regularization $\mathcal{R}$ terms:
\begin{equation}
    \label{eqn:loss-function}
    \mathcal{L}(\hat{z}) = \alpha \mathcal{D}(\hat{z}) + \gamma \mathcal{R}(\hat{z}), \ \ \
    \hat{z} = f(\theta; I, z),
\end{equation}
where $\alpha$ and $\gamma$ are pre-defined positive scalars that are applied \textit{uniformly} to data fidelity and regularization terms to modulate their trade-off.

The model $f$, parameterized by $\theta$, takes an image $I$ and sparse depth $z$, which resides in $\Omega_z \subset \Omega$, as input and produces dense depth $\hat{z} : \Omega \mapsto \real_+$.
To learn depth, we minimize \eqnref{eqn:loss-function} over the entire training dataset.
The data fidelity term $\mathcal{D}$ is designed to penalize the combination of discrepancies (i) between $z$ and its prediction $\hat{z}$ and (ii) between $I$ and its reconstruction $\hat{I}_{\tau}$.
The reconstruction $\hat{I}_{\tau}$ from $I$ is obtained by the following projection equation:
\begin{equation}
    \hat{I}_{\tau}(x) = I_{\tau}\big(\mathrm{p}( \, g_\tau \, K^{-1}
        \begin{bmatrix}
            x \\ 
            1
        \end{bmatrix}
        \hat{z}(x))\big),
\label{eqn:reprojection}
\end{equation}
where $x \in \Omega$, $\tau \in \{ -1, +1 \}$, $g_\tau$ is the relative pose between $I$ and $I_\tau$, $K$ the camera intrinsics, and $\mathrm{p}$ the projection operation.  
There are two main problems in \eqnref{eqn:loss-function}: (i) Because $\mathcal{D}$ is subject to occlusions and disocclusions when registering $I_\tau$ to $I$ and vice versa, occluded and disoccluded regions will yield high reconstruction errors (residuals) and a uniform weighting scheme $\alpha$ will penalize these regions despite the lack of co-visibility.
(ii) Because $\mathcal{R}$ is commonly a local smoothness (e.g. total variation of $\hat{z}$) or a forward-backward consistency term, a uniform weighting scheme $\gamma$ will bias $\hat{z}(x)$ for $x \in \Omega$ to be smooth or consistent with another prediction \textit{without} considering the residuals or correctness of $\hat{z}(x)$, which can cause performance to degrade. 

Hence, neither $\alpha$ nor $\gamma$ should be static, but instead \textit{adapt} to the model and data for each prediction $\hat{z}(x)$. 
As $\alpha$ and $\gamma$ are both related to data fidelity residuals (which evolves throughout training), one must consider the temporal interplay between data fidelity and regularization over the course of optimization. 
Thus, we propose residual-guided adaptive weighting functions $\alpha_\tau(x)$ and $\gamma(x)$, that vary in both space (image domain) and time (optimization step), to determine visibility and regularization.
We combine them into a simple yet effective framework (see \figref{fig:overview}), where their complementary effects (i.e. occluded regions require regularization) can improve baseline unsupervised depth completion algorithms without any additional trainable parameters.

\begin{figure}[t]
    \centering
    \includegraphics[width=1\linewidth]{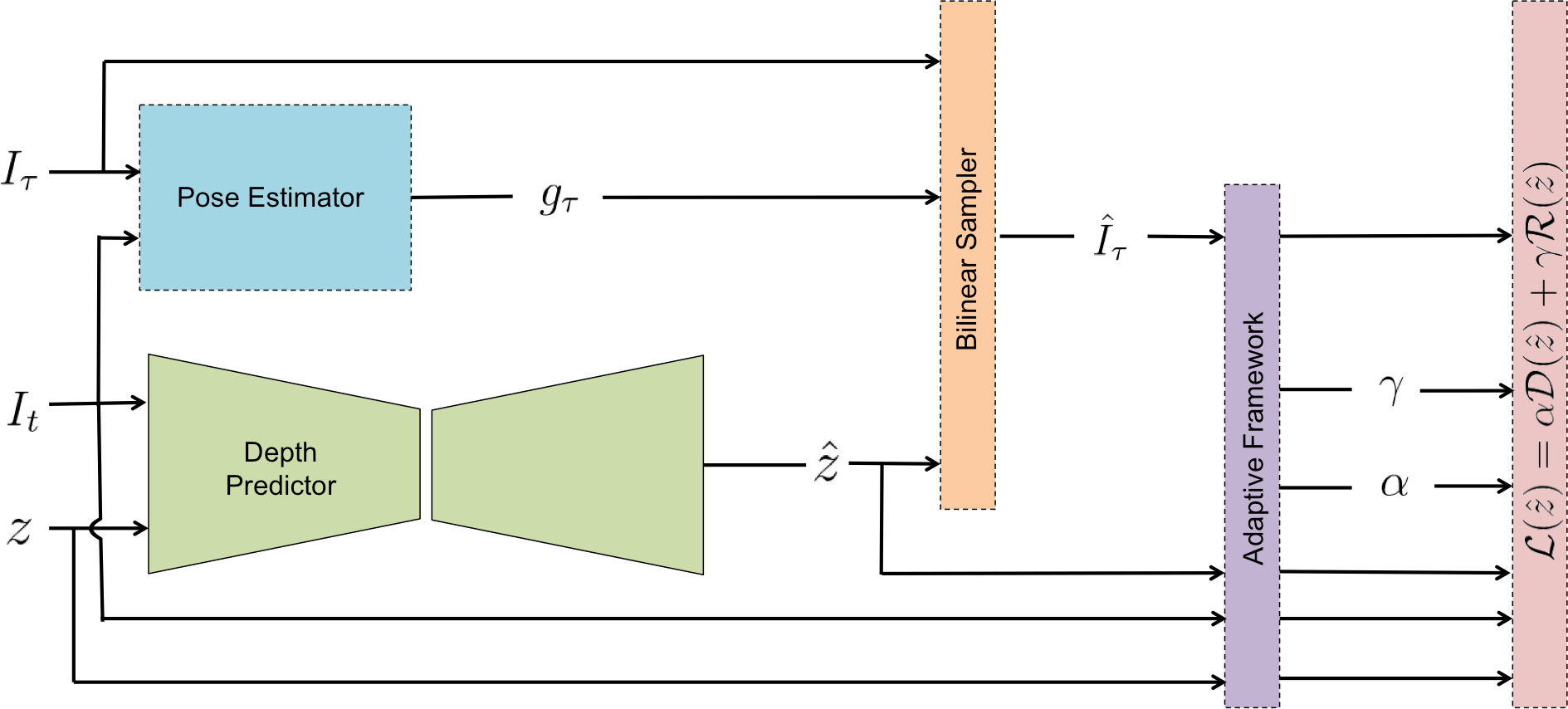}
    \caption{\textit{Diagram of the training pipeline using our framework.} Given the predicted depth $\hat{z}$, sparse depth $z$, image $I_{t}$ and its reconstructions $\hat{I}_{\tau}$, our framework (purple) is comprised of $\alpha$ (consists of $\alpha_{\tau}$ for $\tau \in \{-1, +1\}$) and $\gamma$ for adaptively weighting the data fidelity $\mathcal{D}$ and regularization $\mathcal{R}$ in the loss function (red). Our framework does not require any additional trainable parameters nor additional run-time complexity during inference; the only component required during inference is the depth predictor (green).}
    \vspace{-1.2em}
    \label{fig:overview}
\end{figure}

\section{Determining Visibility Over Time}
\label{sec:determining-visibility-over-time}
Given an image pair ($I$, $I_\tau$) and the depth predictions $\hat{z}(x)$, the reconstruction $\hat{I}_\tau$ suffers from occlusions and disocclusions because $I_\tau$ is captured from a different viewpoint.
A static (uniform weighting) $\alpha$ penalizes all discrepancies between $\hat{I}_\tau$ and $I_\tau$ equally regardless of visibility constraints, i.e. co-visibility, occlusion or disocclusion, and thus \textit{requires} the model to resolve regions that are not co-visible.

Let us consider a scenario where all co-visible correspondences are found, the reconstruction residual will still be non-zero and hence the gradients will continue to update the model parameters $\theta$ to find unresolvable correspondences up to the allowed regularization, causing the model to \textit{move away} from the desired solution.
One may discount occlusions and disocclusions with a binary mask based on a fixed threshold (i.e. in traditional SFM, stereo). This is applicable at convergence, when all correspondences have been found. However, at early time steps, predictions are largely random and hence will yield high residuals. Thresholding would discount the training signal and in turn \textit{impede learning}.
Hence, an adaptive weighting scheme $\alpha_\tau \in [0, 1]$ for ($I, I_\tau$) should weight all pixels equally at the early stages of training. As the model becomes more confident in the correspondences found over the course of training, $\alpha_\tau \rightarrow 0$ for regions with high residuals, gradually discounting the errors.

\subsection{Residual Function}
We begin with a simple residual function as a measure for determining  whether a pixel is co-visible, or occluded or disoccluded. Assuming images with intensity range of $[0, 1]$:
\begin{equation}
    \delta_{\tau}(x) = |I(x)-\hat{I}_{\tau}(x)| \text{ for } x \in \Omega
\label{eqn:delta-image}
\end{equation}
measures the discrepancy between $I$, and its reconstruction $\hat{I}_{\tau}$ (photometric error). 
Note: $\delta_\tau$ can be replaced by a more sophisticated measure such as SSIM~\cite{wang2004image}, but we aim to demonstrate the effectiveness of our proposed scheme with a simple one.
We then normalize the residual $\delta_{\tau}$ to have a zero-mean distribution with unit variance.
\begin{equation}
    \mu_{\tau} = \frac{1}{|\Omega|} \sum_{x \in \Omega} \delta_{\tau}(x), \ \ \
    \sigma_{\tau}^2 = \frac{1}{|\Omega|} \sum_{x \in \Omega} (\delta_{\tau}(x) - \mu_{\tau})^2, 
\end{equation}
\begin{equation}
    \rho_{\tau}(x) = \frac{\delta_{\tau}(x) - \mu_{\tau}}{\sqrt{\sigma_{\tau}^2 + \epsilon}},
    \label{eqn:rho}
\end{equation}

where $x \in \Omega$ and $\epsilon$ is a small positive scalar used for numerical stability.
In the next section, we will use $\rho_{\tau}$  as a cue to determine if a pixel is co-visible, or occluded or disoccluded by constructing a soft visibility mask $\alpha_{\tau}$ that evolves over training time.

\begin{figure}[t]
    \centering
    \includegraphics[width=0.75\linewidth]{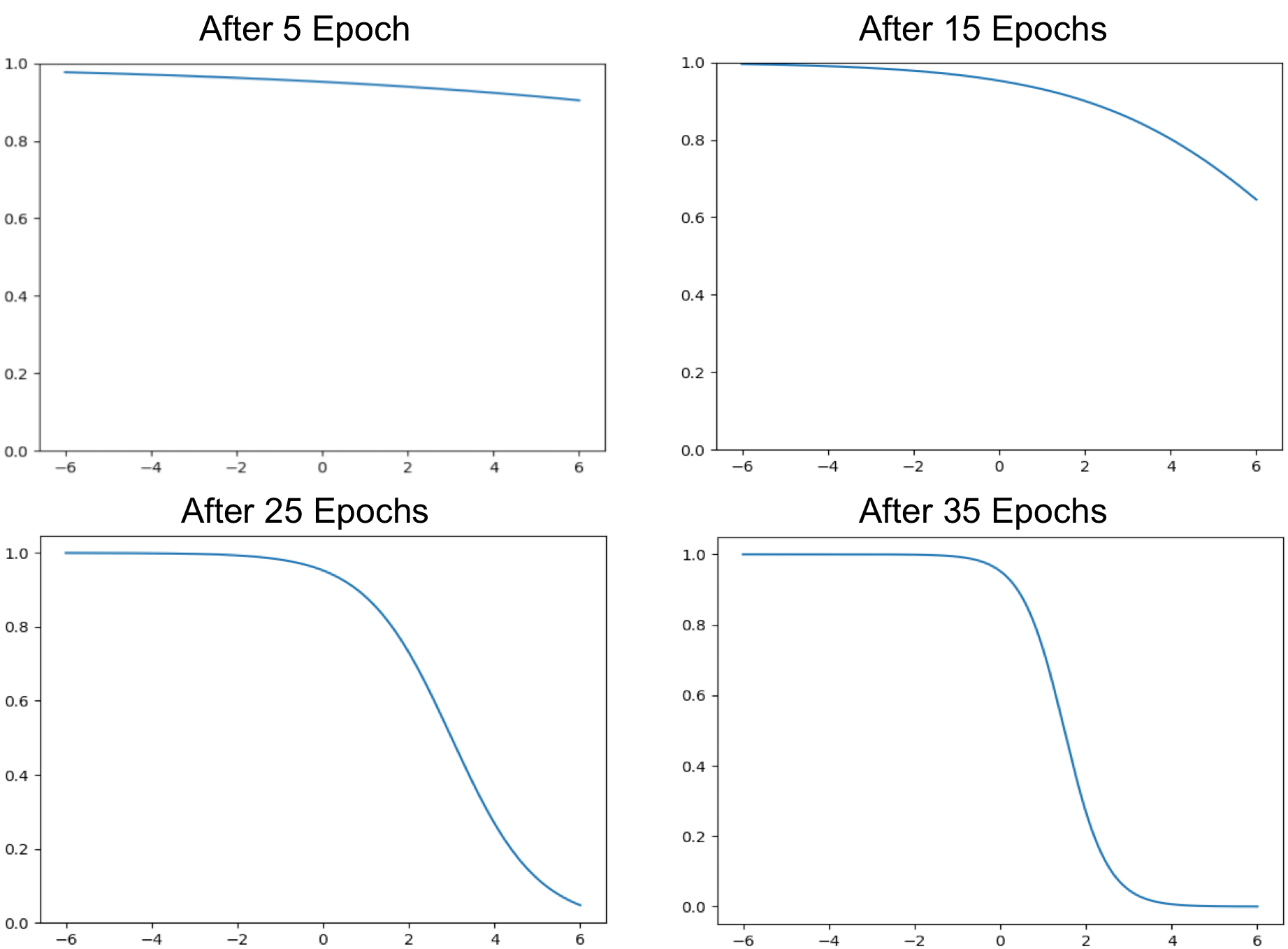}
    \caption{\textit{The shape of $\alpha_\tau$ from mean residual $\mu_\tau$ sampled after the $5^{th}$, $15^{th}$, $25^{nd}$ and $35^{th}$ epoch}. x-axis denotes $\rho_\tau(x)$ and y-axis denotes the value of $\alpha_\tau$. $\alpha_\tau$ begins at a flat curve close to 1 and over time sharpens into a flipped sigmoid. Binary thresholding is a special case of our approach.}
    \vspace{-1.2em}
    \label{fig:alpha-function-over-time}
\end{figure}

\subsection{Discounting Occlusions and Disocclusions}
\label{sec:discounting-occlusions-disocclusions}
The weighting function $\alpha_{\tau}$ assigns the probability of co-visibility between $I$ and $\hat{I}_{\tau}$ for each pixel by adaptively adjusting a flipped sigmoid based on $\rho_{\tau}$ for every time step (\figref{fig:alpha-function-over-time}). Co-visible pixels will have a higher weight, while occluded or dis-occluded pixels will have a lower weight:
\begin{equation}
    \alpha_{\tau}(x) = 1 - \frac{1}{1 + \exp(-(a\rho_{\tau}(x)-b))},
    \label{eqn:alpha}
\end{equation}
where $a > 0$ controls the curvature (steepness) of the sigmoid and $b \geq 0$ the shift.
To enable adaptation over training time, we vary $a$ and $b$ based on the mean residual $\mu_{\tau} \in [0, 1]$.
The steepness parameter $a$ is designed to gradually increase over training as the overall residual decreases:
\begin{equation}
    a = \frac{a_0}{\mu_{\tau}+\epsilon}
    \label{eqn:a}
\end{equation}
where $a_0$ is a positive scalar based on the range of image intensity.
As we are unsure of the correspondences during the early stages of training, $\alpha_{\tau}$ should be uniform over the spatial domain $\Omega$, which occurs as $a \rightarrow 0$. Towards convergence, $\alpha_{\tau}$ takes on the shape of a flipped sigmoid to discount occlusions and disocclusions. Hence, we let $a$ be inversely proportional to the mean residual $\mu_{\tau}$ and we choose $a_0$ to be close to 0. At the start of training, $\mu_{\tau}$ is large (making $a$ small) and $\alpha_{\tau}$ tends to a flat curve. As we converge, $\mu_{\tau}~\rightarrow~0$, making $a$ large and giving $\alpha_{\tau}$ sharper curvature. 

Similarly, we also allow the shift parameter $b$ of $\alpha_{\tau}$ to vary over training time by making it a function of $\mu_{\tau}$:
\begin{equation}
  b = b_0(1 - \cos(\pi \, \mu_{\tau})),
  \label{eqn:b}
\end{equation}
where $b_0$ is a positive constant used as the upper bound of the shift and $\mu_{\tau} \in [0, 1]$ leading to $b \in [0, 2 b_0]$, following a cosine decay rate.
At the early time steps, $\mu_{\tau}$ is large, and thus $b \rightarrow 2b_0$ causing $\alpha_{\tau}$ to tend to 1. As residuals decrease over training time, $b \rightarrow 0$, resulting in $\alpha_{\tau}$ being a centered flipped sigmoid function. 

\begin{figure}[t]
    \centering
    \includegraphics[width=1\linewidth]{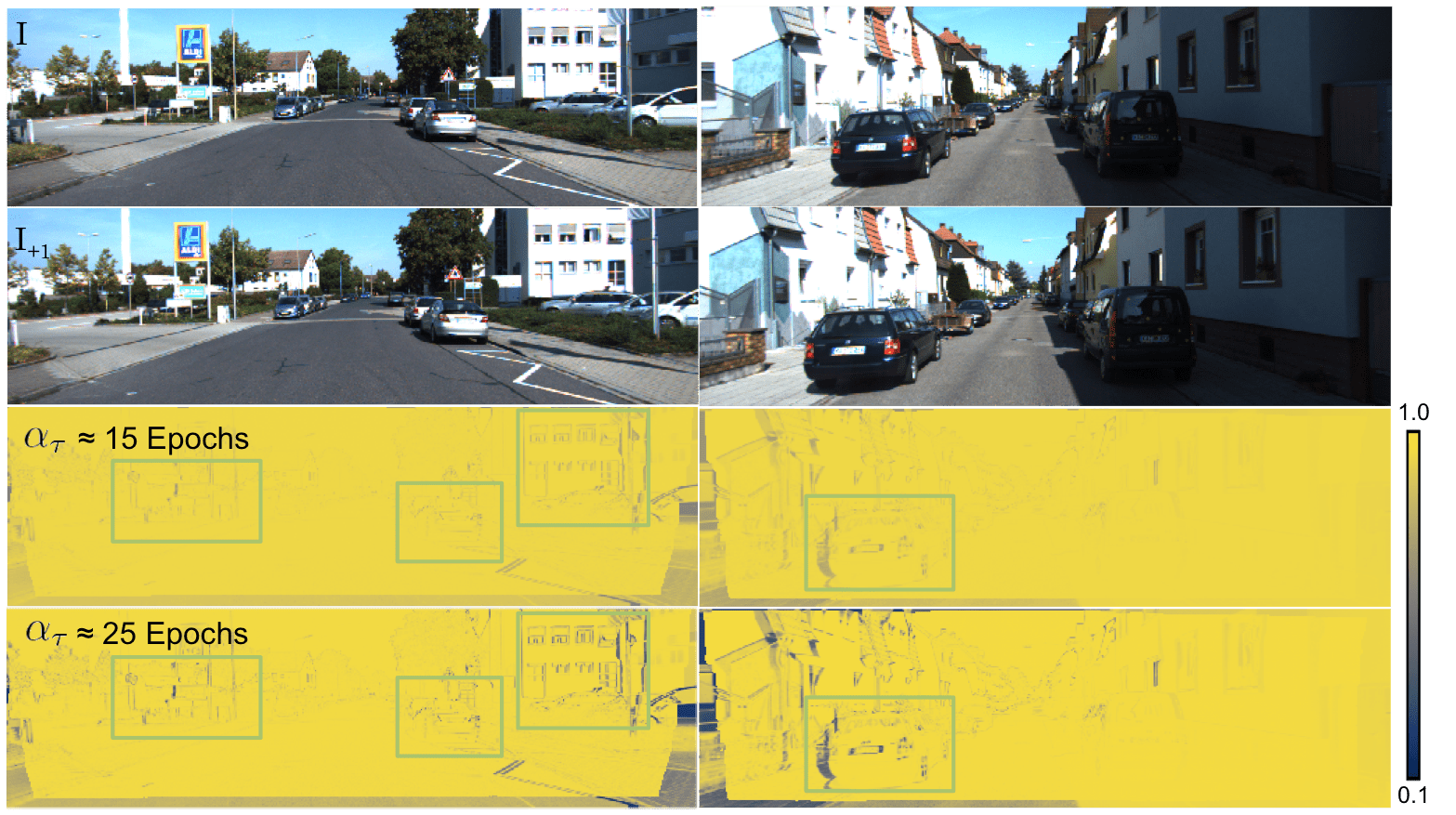}
    \caption{\textit{$\alpha_\tau$ over training time}. $\alpha_\tau$ varies spatially and over training time, and reduces weight of occlusions and disocclusions regions (e.g. the borders of the image and regions highlighted in green) as we become more confident in correspondences between $I$ and $I_{+1}$.}
    \vspace{-1em}
    \label{fig:alpha-example}
\end{figure}

By making $a$ and $b$ a function of the mean residual $\mu_{\tau}$, the weighting function $\alpha_{\tau}$ becomes an annealing process to detect occlusions or disocclusions. Because $\alpha_{\tau}$ is modulated by both the local (per-pixel) residual as well as the mean residual (generally decreases throughout training), $\alpha_{\tau}$ will vary over both the image spatial domain and training time.
For every $x \in \Omega$, $\alpha_{\tau}(x) \approx 1$ at the early stage of the training, whereas $\alpha_{\tau}(x)$ approaches either 0 or 1 towards the convergence of the training.
We note that the binary mask produced by thresholding is a special case of our method with specific $a$ and $b$.
We construct \figref{fig:alpha-function-over-time} by sampling $\mu_{\tau}$ over the course of training to illustrates how $\alpha_{\tau}$ is guided by mean residual and varies over training time. 
\figref{fig:alpha-example} shows $\alpha_\tau$ as an image. 
The co-visible pixels (yellow) are assigned higher weight; whereas, the occluded and disoccluded ones (blue) are assigned lower weight -- as we train, the weight of those regions decreases as we are more confident in our predictions.
In the data fidelity term $\mathcal{D}$, one can apply $\alpha_{\tau}$ to the photometric error $\mathcal{D}_{ph}$ simply by:
\begin{equation}
    \mathcal{D}_{ph}(\hat{z}) = \frac{1}{|\Omega|} \sum_{x \in \Omega} \alpha_{\tau}(x)|I(x)-\hat{I}_{\tau}(x)|.
\label{eqn:example-alpha}
\end{equation}

\section{Adaptive Regularization}
Regularization is typically imposed uniformly over the prediction to make the depth completion problem well-posed.
For instance, a local smoothness term assumes a smooth transition in $\hat{z}(x)$ and penalizes discontinuities, but does not account for object boundaries where depth discontinuities generally occur. Hence, uniformly imposing regularity may lead to an undesirably biased model (e.g. over-smoothing). 
To allow discontinuities along object boundaries, previous works, including, but not limited to~\cite{ma2019self,shivakumar2019dfusenet,wong2020unsupervised}, ``adapt'' to the data by weighting $\mathcal{R}(x)$ based on the image gradients $\nabla I(x)$ -- reducing $\gamma(x)$, the regularization parameter, in textured regions.
However, $\gamma(x)$  is \textit{still static} with respect to the image (same weights for the same image).
Also, this does not consider \textit{residuals} where regularization not only propagates the incorrect solution, but also restricts the model from exploring the solution space (i.e. predicting large disparities). This also holds for other regularizers, such as temporal consistency; enforcing consistency with incorrect predictions only introduces more errors. 

Hence, $\gamma \in [0, 1]$ should adaptively imposes regularization based on residuals from both image and sparse depth. 
$\gamma(x)$ follows two simple principles for a given $\hat{z}(x)$:  (i) the higher the residual, the lower the regularity. This not only lowers the influence of incorrect predictions in a local neighborhood, but also gives a model the flexibility to maximize its fitness to data. (ii) the earlier the time step, the lower the regularity. The local residual at early steps can be low depending on initialization, applying regularization effectively limits the scope of the solution space. Hence, small amounts of regularity should be imposed to allow the model to explore. To illustrate these seemingly counter-intuitive principles, let's consider stereo matching. One can predicted disparity up to the amount allowable by regularization. If $\gamma(x)$ is large, then one cannot find long range correspondences; hence, we want to reduce $\gamma(x)$. Once the correspondence is found, we can leverage the correct prediction to inform its neighbors' predictions (e.g. local smoothness).

\subsection{Residual Functions}
We will reuse the image reconstruction residual (\eqnref{eqn:delta-image}) as our adaptation signal from an image. As we assume that there are two temporally adjacent frames, $I_\tau$ for $\tau~\in~\{-1, +1\}$, there exists two reconstructions of $I_\tau$ to guide $\gamma$. Following our first principle to apply regularization when residual is low, for each $x \in \Omega$, we choose the minimum residual of the two reconstructions:
\begin{equation}
    \delta_i(x) = \displaystyle \min_\tau(\delta_{\tau}(x)) \text{ for } x \in \Omega.
\label{eqn:delta-multi-image}
\end{equation}

To obtain an adaptation signal from depth input, we consider the sparse depth reconstruction residual:
\begin{equation}
    \delta_z(x) = 
\begin{cases}
    |\hat{z}(x)-z(x)|, & \text{ if } x \in \Omega_z, \\
    0,         & \text{ if } x \in \Omega \backslash \Omega_z.
\end{cases}
\label{eqn:delta-sparse-depth}
\end{equation}

Next, we will use $\delta_i$ and $\delta_z$ to construct $\gamma_i$ (from image) and $\gamma_z$ (from sparse depth), and combine them to form our adaptive regularization weighting scheme $\gamma$.

\begin{figure}[t]
    \centering
    \includegraphics[width=1\linewidth]{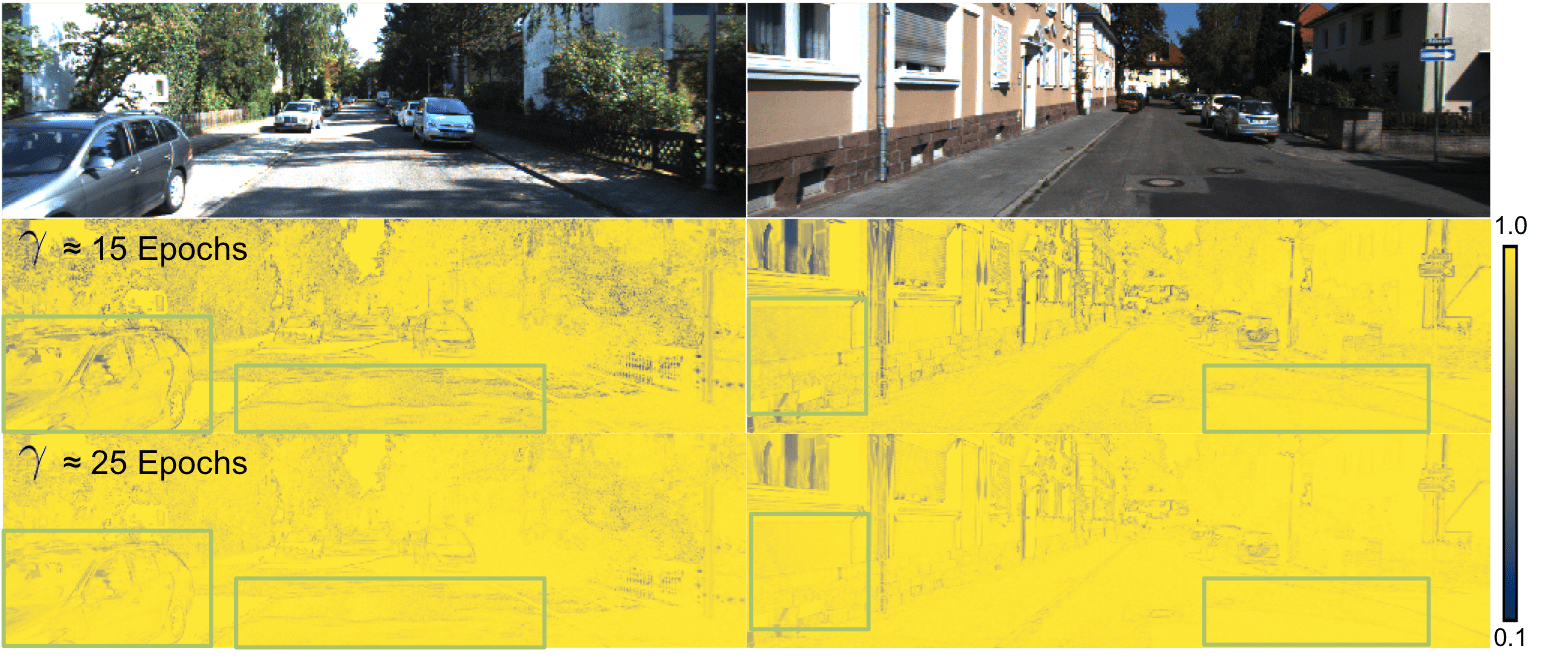}
    \caption{\textit{$\gamma$ over training time}. $\gamma$ is low during early stages of training and increases over time as mean residuals decrease. Regions of small residuals due to noise and slight illumination change (highlighted) are gradually regularized. For a local smoothness term, $\gamma$ first allow the model to search for better correspondences and gradually impose smoothness.}
    \vspace{-1em}
    \label{fig:gamma-example}
\end{figure}

\subsection{Image and Sparse Depth as Guidance}
\label{sec:fusing-image-depth-to-determine-regularity}
To realize our second principle of having a small $\gamma(x)$ at early time steps, we note a keen observation: while some local residuals may be small, the mean residual will be large and will gradually decrease over the course of optimization -- making it a good proxy for training time. Hence, $\gamma(x)$ should be inversely proportional to the mean residual. First, we model $\gamma_i$, adaptive weights guided by image residuals, with a negative exponential function: 
\begin{align}
    \gamma_i(x) &= \exp(-c_i \, \mu_i \, \delta_i(x)) \text{ for } x \in \Omega
    \label{eqn:gamma-image}\\
    \mu_i &= \displaystyle \frac{1}{|\Omega|} \sum_{x \in \Omega} \delta_i(x) 
    \label{eqn:mu-image}
\end{align}

where $c_i$ is a positive scalar based on the range of image intensities. Similarly, we also construct $\gamma_z$, the set of adaptive weights from sparse depth residuals:
\begin{align}
    \gamma_z(x) &= \exp(-c_z \, \mu_z \, \delta_z(x)) \text{ for } x \in \Omega_z
    \label{eqn:gamma-sparse-depth} \\
    \mu_z &= \frac{1}{|\Omega_z|} \sum_{x \in \Omega_z} \delta_z(x) \label{eqn:mu-sparse-depth}
\end{align}

where $c_z$ is a positive scalar based on the range of depth measurements. 
Both $\gamma_i$ and $\gamma_z$ are modulated by their respective local and mean residuals. At early steps, both are low and increase over time, except where $\hat{z}(x)$ yield high residuals. We note that modulating $\gamma_i$ and $\gamma_z$ with their mean residuals as a proxy of training time is more stable than using discrete training steps, which have no upper bound. If $\gamma_i$ and $\gamma_z$ directly depend on training steps, then they may modify the model even after convergence, and introduce instability. In contrast, the mean residual stays approximately constant at convergence and $\gamma_i$ and $\gamma_z$ will like-wise be stable.

Lastly, as noted by previous works, sparse depth and image may conflict due to noise in depth sensor, and illumination changes in images. To combine $\gamma_i$ and $\gamma_z$, we assume depth measurements (when available) are more reliable and choose $\gamma_z$ over $\gamma_i$, yielding the final adaptive weights:
\begin{equation}
    \gamma(x) = 
\begin{cases}
    \gamma_z(x), & \text{ if } x \in \Omega_z, \\
    \gamma_i(x), & \text{ if } x \in \Omega \backslash \Omega_z.
\end{cases}
\label{eqn:gamma}
\end{equation}

The behavior of $\gamma$ is similar to anisotrophic diffusion at convergence since the regions of high residuals will be occlusion or disocclusions, which generally occurs across object boundaries (see \figref{fig:gamma-example}). However, unlike \secref{sec:discounting-occlusions-disocclusions}, we chose a negative exponential over a sigmoid function because the negative exponential is less aggressive at convergence. Recall that $\alpha_{\tau}$ approaches a binary mask, but we still need some regularity since the problem is ill-posed. 

Assuming local smoothness as  $\mathcal{R}$, one can apply $\gamma$ by:
\begin{equation}
    \mathcal{R}(\hat{z}) = \frac{1}{|\Omega|} \sum_{x \in \Omega} \gamma(x) ||\nabla \hat{z}(x)||^2
\end{equation}

Together, $\alpha_\tau$ and $\gamma$ \textit{complement each other}. During early time steps, $\alpha_\tau$ is high and $\gamma$ is low, allowing the model to explore the solution space for better correspondences. As residuals decrease over time, $\alpha_\tau$ discovers occlusions or disocclusions and discounts them. This is precisely when we need regularity and consequently $\gamma$ increases (see \figref{fig:alpha-example}, \ref{fig:gamma-example}).

We note that $\alpha_\tau$ and $\gamma$ are general and can be constructed with a stereo pair as well. In this case, there is only one reconstruction from a stereo-counterpart, $\delta_i(x)$ is simply the reconstruction residual (\eqnref{eqn:delta-image}) instead of the minimum residual from multiple views (\eqnref{eqn:delta-multi-image}). To show that our framework is applicable to both stereo and monocular training paradigms, we use \cite{shivakumar2019dfusenet} as a baseline and construct $\alpha_\tau$ and $\gamma_i$ using stereo pairs (see \tabref{tab:kitti-validation-set}).

\begin{figure*}[h]
    \centering
    \includegraphics[width=1\linewidth]{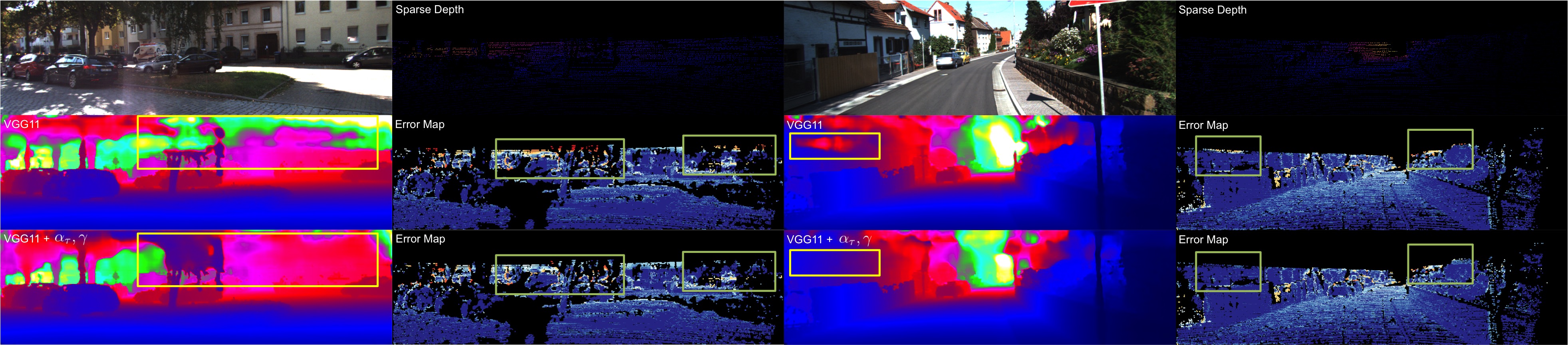}
    \caption{\textit{KITTI depth completion test set}. VGG11 completely missed the building, and tree on the left and the wall on the right (highlighted in yellow). By considering the fitness of the model to the data, our framework enables VGG11 to recover all of them. Green boxes highlight error regions for comparison.}
    \label{fig:kitti-test-set}
\end{figure*}

\section{Implementation Details}
All models using our framework are trained from scratch. Our framework consists of computationally cheap operations and only increases training time by $\approx 2.2\%$, and incurs no additional parameters or inference time. 

\textit{Hyper-parameters}: $\alpha_\tau$ and $\gamma$ are set based on the range of input. Image intensities are scaled between 0 and 1 for both KITTI and VOID. Depth ranges from 1m to 100m in KITTI and 0.1m to 10m in VOID. We choose $a_0 = 0.10$, $b_0 = 4.0$ and $\epsilon = 10^{-8}$ for $\alpha_\tau$. The same $a_0$ and $b_0$ are used for both $\alpha_\tau$. We set $c_i = 1.0$, $c_z = 0.01$ for $\gamma$ to adjust for the difference in magnitude between image and depth values. The same hyper-parameters are used for all methods for both KITTI and VOID except for $c_i$, which we set to $0.70$ (less aggressive weighting) for VOID since indoor scenes contains more textureless surfaces and requiring more regularity. For the same reason, to train \cite{ma2019self} on VOID, we set the weight of their smoothness term to $1.0$ ($10\times$ their proposed weight). 

\begin{table}[t]
\centering
\caption{Error metrics}
\setlength\tabcolsep{22pt}
\begin{tabular}{l l}
\midrule
    Metric & Definition \\ \midrule
    MAE &$\frac{1}{|\Omega|} \sum_{x\in\Omega} |\hat z(x) - z_{gt}(x)|$ \\
    RMSE & $\big(\frac{1}{|\Omega|}\sum_{x\in\Omega}|\hat z(x) - z_{gt}(x)|^2 \big)^{1/2}$ \\
    iMAE & $\frac{1}{|\Omega|} \sum_{x\in\Omega} |1/ \hat z(x) - 1/z_{gt}(x)|$ \\
    iRMSE& $\big(\frac{1}{|\Omega|}\sum_{x\in\Omega}|1 / \hat z(x) - 1/z_{gt}(x)|^2\big)^{1/2}$ \\ \midrule
\end{tabular}
\begin{tablenotes}
    Error metrics used in \tabref{tab:kitti-validation-set}, \ref{tab:void-test-set}. $z_{gt}$ denotes the ground truth.
\end{tablenotes}
\label{tab:error-metrics}
\end{table}

\section{Experiments and Results}
\label{sec:experiments-results}
We applied our adaptive framework ($\alpha_\tau$ and $\gamma$) to recent unsupervised depth completion methods and evaluate the relative improvements on the KITTI \cite{uhrig2017sparsity} in \secref{sec:kitti-depth-completion-benchmark} (outdoors) and VOID \cite{wong2020unsupervised} in \secref{sec:void-depth-completion-benchmark} (indoors).

\begin{table}[t]
\centering
\setlength\tabcolsep{6.8pt}
\caption{Quantitative results on KITTI validation set}
\begin{tabular}{l c c c c}
    \midrule 
    Method & MAE & RMSE & iMAE & iRMSE \\ 
    \midrule
    Ma \cite{ma2019self}
    & 358.92 & 1384.85 & 1.60 & 4.32   \\ 
    \midrule 
    Ma \cite{ma2019self} + $\alpha_\tau, \gamma$
    & \textbf{332.54} & \textbf{1301.42} & \textbf{1.43} & \textbf{4.01} \\ 
    \midrule
    Shivakumar \cite{shivakumar2019dfusenet}
    & 396.43 & 1285.79 & 1.37 & 4.05 \\ 
    \midrule
    Shivakumar \cite{shivakumar2019dfusenet} +  $\alpha_\tau, \gamma$
    & \textbf{346.18} & \textbf{1231.06} & \textbf{1.31} & \textbf{3.84} \\ 
    \midrule
    VGG8 \cite{wong2020unsupervised}  
    & 308.81 & 1230.85 & 1.29 & 3.84 \\ 
    \midrule
    VGG8 \cite{wong2020unsupervised}  +  $\alpha_\tau, \gamma$
    & \textbf{298.89} & \textbf{1189.43} & \textbf{1.18} & \textbf{3.64} \\ 
    \midrule
    VGG11 \cite{wong2020unsupervised} 
    & 305.06 & 1239.06 & 1.21 & 3.71 \\ 
    \midrule
    VGG11 \cite{wong2020unsupervised} + $\alpha_\tau, \gamma$
    & \textbf{291.57} & \textbf{1186.07} & \textbf{1.16} & \textbf{3.58} \\ 
    \midrule
    \end{tabular}
    \begin{tablenotes}
        Results of \cite{ma2019self,wong2020unsupervised} are taken from their papers. Results of \cite{shivakumar2019dfusenet} were not available; hence, we train \cite{shivakumar2019dfusenet} from scratch. Our approach (entries with + $\alpha_\tau, \gamma$) consistently improves all methods across all metrics.
    \end{tablenotes}
\label{tab:kitti-validation-set}
\end{table}

\begin{figure*}[ht]
    \centering
    \includegraphics[width=1\linewidth]{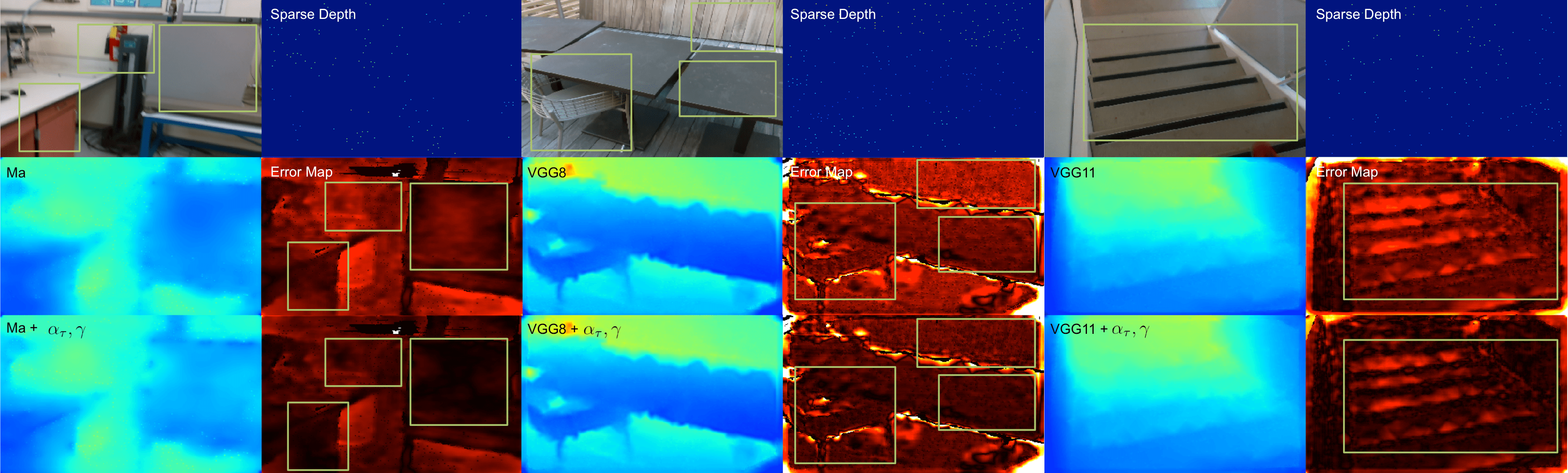}
    \caption{\textit{VOID depth completion test set}. We apply our framework to \cite{ma2019self}, VGG8, and VGG11 of \cite{wong2020unsupervised}. Our approach consistently improves the overall scene. While depth maps may look similar, the error map of each method using our framework (+ $\alpha_\tau, \gamma$) is a shade of red darker (lower error). For example, We observe large improvements in smooth surfaces (tables, walls, staircases) Green boxes highlight regions for comparison.}
    \label{fig:void-test-set}
\end{figure*}

\subsection{KITTI Unsupervised Benchmark}
\label{sec:kitti-depth-completion-benchmark}
KITTI provides $\approx 80,000$  synchronized stereo pairs and sparse depth maps of $\approx 1242 \times 375$ resolution for outdoor driving scenes. The sparse depth maps are captured by a Velodyne lidar sensor ($\approx 5\%$ density) and projected onto the image frame. The ground-truth depth map is created by accumulating the neighbouring 11 raw lidar scans, with dense depth corresponding to the lower $30\%$ of the images. 

We apply our framework to \cite{ma2019self}, \cite{shivakumar2019dfusenet}, and  VGG8, and VGG11 of \cite{wong2020unsupervised} and evaluate them on the KITTI validation set in \tabref{tab:kitti-validation-set} using error metrics in \tabref{tab:error-metrics}. Due to the limit of one entry per method on the KITTI online test benchmark, we chose to show the relative improvements on the validation set -- before and after applying $\alpha_\tau$ and $\gamma$. The results listed are taken directly from their papers except for \cite{shivakumar2019dfusenet}, which were not reported. We trained their model from scratch in \tabref{tab:kitti-validation-set}. While we primarily focus on the monocular training, we include \cite{shivakumar2019dfusenet} to show that our framework can also be applied to and improve methods using stereo training. 

\tabref{tab:kitti-validation-set} shows that our framework consistently improves all methods across all metrics. While our method can be used to improve both existing and yet-to-be-developed methods, the real test is whether it can boost an underperforming method over the state of the art. Hence, a \textbf{key comparison} is between VGG8, VGG8 + $\alpha_\tau, \gamma$ and VGG11. Indeed, our framework improves an inferior method, VGG8, over the state-of-the-art VGG11 across all metrics on the KITTI depth completion validation set as well as the official online KITTI depth completion test set (Table IV, Supp. Mat.) to achieve the state of the art on unsupervised depth completion. \figref{fig:kitti-test-set} shows that our framework can help VGG8 more correctly recover the scene. We note that the performance boost is \textit{almost free} ($\approx +2.2\%$ in training time) -- there is no additional parameters, pre- or post-processing, nor increase in inference time. The gain is solely from guiding the learning (optimization) process via adaptively weighting their objective function.

\begin{table}[t]
    \centering
    \setlength\tabcolsep{8.6pt}
    \caption{VOID test set and ablation study on $\alpha_\tau$ and $\gamma$}
    \begin{tabular}{l c c c c}
        \midrule 
        Method & MAE & RMSE & iMAE & iRMSE \\ \midrule
        Ma \cite{ma2019self}
        & 178.85 & 243.84 & 80.12 & 107.69 \\ \midrule 
        Ma \cite{ma2019self} + $\alpha_\tau, \gamma$
        & \textbf{154.48} & \textbf{220.63} & \textbf{64.68} & \textbf{91.77} \\ \midrule
        VGG8 \cite{wong2020unsupervised} 
        & 98.45 & 169.17 & 57.22 & 115.33 \\ \midrule
        VGG8 \cite{wong2020unsupervised}   +  $\alpha_\tau, \gamma$
        & \textbf{86.25} & \textbf{153.05} & \textbf{49.26} & \textbf{94.74} \\ \midrule
        VGG11 \cite{wong2020unsupervised}  
        & 85.05 & 169.79 & 48.92 & 104.02 \\ \midrule
        VGG11 \cite{wong2020unsupervised}  + $\alpha_\tau$
        & 83.24 & 139.52 & 47.51 & 83.69 \\ \midrule
        VGG11 \cite{wong2020unsupervised}  + $\gamma$
        & \textbf{78.20} & 140.86 & 45.41 & 85.20 \\ \midrule
        VGG11 \cite{wong2020unsupervised}  + $\alpha_\tau, \gamma$ 
        & 78.79 & \textbf{135.93} & \textbf{ 43.62} & \textbf{78.22}  \\ \midrule
    \end{tabular}
    \begin{tablenotes}
        Our framework (+ $\alpha_\tau, \gamma$) consistently improves all methods across all metrics. $\alpha_\tau$ and $\gamma$ provide complementary benefits. The ablation study (last 4 rows) on VGG11 shows that $\gamma$ improves MAE and iMAE, and $\alpha_\tau$, RMSE and iRMSE.  When used together, they achieve the best results..
    \end{tablenotes}
\label{tab:void-test-set}
\end{table}

\subsection{VOID Unsupervised Benchmark}
\label{sec:void-depth-completion-benchmark}
VOID provides $\approx 47,000$ synchronized images and sparse depth maps of $640 \times 480$  resolution of indoor scenes. Sparse depth ($\approx 1500$ points, covering $\approx 0.5\%$ of the image) are the set of features tracked by XIVO \cite{fei2019geo}. The ground-truth depth maps are dense and are acquired by active stereo. The testing set contains $800$ frames.

We apply our framework on \cite{ma2019self}, VGG8 and VGG11. Since \cite{ma2019self} and VGG8 did not report results on VOID, we trained their models from scratch. Note: \cite{shivakumar2019dfusenet} requires stereo pairs for training, so we cannot train their model on VOID. 

VOID consists of indoor scenes with many textureless surfaces (e.g. walls, cabinets) and non-trivial 6 degrees of freedom motion. Hence, (i) regularization is even more important as the data fidelity term does not provide useful local information. This is where $\gamma$ is helpful. By adjusting the regularity based on residuals, $\gamma$ allows the model to find correspondences first, then impose regularization. Moreover, (ii) due to the large motion, occlusions and disocclusions can easily cause the model to leave a desirable local minimum. $\alpha_\tau$ mitigates their impact by discounting them over time. The effectiveness of our framework can be seen in \tabref{tab:void-test-set} and \figref{fig:void-test-set}, where we improved all methods by large margins across all metrics. We hypothesize the large gain in iMAE and iRMSE metrics may be due to the low density of depth measurements. Hence, the model must rely heavily on the signal from the image, which is guided by $\alpha_\tau$ and $\gamma$. 

As an ablation study, we examine $\alpha_\tau$ and $\gamma$ (\tabref{tab:void-test-set}) individually on VGG11 and find that both provide complementary benefits. $\gamma$ provides more improvements to MAE and iMAE while $\alpha_\tau$ improves RMSE and iRMSE. Note: the MAE improvement from $\gamma$ (last 2 rows) is comparable to our full model. This is because $\alpha_\tau$ reduces outliers (as measured by RMSE metrics) caused by occlusions and disocclusions while $\gamma$ improves the overall accuracy of the scene (as measured by MAE metrics) through regularization. 

\begin{table}[!t]
    \centering
    \setlength\tabcolsep{10.1pt}
    \caption{Ablation study of various density levels on VOID test set}
    \begin{tabular}{l c c c c}
        \midrule 
        Method & MAE & RMSE & iMAE & iRMSE  \\ \midrule
        & \multicolumn{4}{c}{$\approx$0.50\% density} \\ \midrule
        VGG11 \cite{wong2020unsupervised}
        & 85.05 & 169.79 & 48.92 & 104.02  \\ \midrule
        VGG11 + $\alpha_{\tau}$, $\gamma$ 
        & \textbf{78.79} & \textbf{135.93} & \textbf{43.62} & \textbf{78.22}  \\ \midrule
        & \multicolumn{4}{c}{$\approx$0.15\% density} \\ \midrule
        VGG11 \cite{wong2020unsupervised} 
        & 124.11 & 217.43 & 66.95 & 121.23 \\ \midrule
        VGG11 + $\alpha_{\tau}$, $\gamma$ 
        & \textbf{112.31} & \textbf{188.60} & \textbf{59.47} & \textbf{101.26}
        \\ \midrule
        & \multicolumn{4}{c}{$\approx$0.05\% density} \\ \midrule
        VGG11 \cite{wong2020unsupervised}
        & 179.66 & 281.09 & 95.27 & 151.66 \\ \midrule
        VGG11 + $\alpha_{\tau}$, $\gamma$ 
        & \textbf{155.01} & \textbf{262.54} & \textbf{83.55} & \textbf{140.98} \\ \midrule
    \end{tabular}
    \begin{tablenotes}
        The percent density levels correspond to roughly 1500, 500, and 150 points, respectively. By applying our framework to VGG11 \cite{wong2020unsupervised}, we improve their model across all metrics and consistently across all density levels. These are the scenarios where unsupervised depth completion methods must rely on the image -- due to the lack of sparse points. They are also the scenarios where our framework can provide improvements, especially for indoor scenarios. Thus, even at $\approx$0.05\% density, we still boost performance.
    \end{tablenotes}
\label{tab:void-density-ablation}
\end{table}

To evaluate the effect of different density levels, we provide an ablation study on the VOID \cite{wong2020unsupervised} dataset, which provides three levels: $\approx$0.50\%, $\approx$0.15\% and $\approx$0.05\% of the image space -- each of these densities corresponds to roughly 1500, 500, and 150 points. In \tabref{tab:void-density-ablation}, we show the results of VGG11 \cite{wong2020unsupervised}, directly taken from their paper, and the results of VGG11 trained with our framework. We observe consistent improve across all metrics and across all density levels. As the density of the input sparse depth decreases, one \textit{must} rely on the image even more. For indoor, this becomes difficult as surfaces are commonly textureless and motion is more challenging (causing occlusions and dis-occlusions). 

This is precisely where our method can provide improvements. Our framework produces a soft visibility mask $\alpha_\tau$ to deal with occlusions and dis-occlusions and $\gamma$ to determine the strength of regularization, which, in this setting, generally involves local smoothness and forward-backward consistencies. $\alpha_\tau$ discounts occlusion and dis-occlusions over time so that the model does not get driven out of a desirable local minimum due to unresolvable residuals. In the case where correspondences are not found, $\gamma$ allows the model enough flexibility to search for long-range matches (as opposed to uniform weight, which may restrict the model to shorter distances in the image space depending on selected scalar). Once a correspondence is found, $\gamma$ increases regularization and propagates the solution to its neighbors, which directly impacts textureless regions, occluded and dis-occluded regions, and prevents over-smoothing. Together, $\alpha_\tau$ and $\gamma$ play complementary roles by discounting residuals at occluded and dis-occluded regions while propagating depth values from co-visible regions to those locations.  

\section{Discussion}
We have provided a general residual-driven framework for determining co-visibility and the degree of regularization over the optimization process. While our framework improves unsupervised depth completion methods without compromising run-time, it does require tuning several parameters depending on the range of sensors and environment. We use simple measure of residual and do not consider sparse depth in $\alpha_\tau$. We also assume depth measurements are reliable than images when constructing $\gamma$. In reality, both camera and depth sensors have failure modes. Perhaps considering hardware uncertainty can better combine the two. We leave this for future work. There is a long road ahead, but we hope that our framework can motivate further studies in adaptive learning schemes in sensor fusion.

\bibliographystyle{ieee}
\bibliography{condensebib}

\vspace{2em}

\begin{center}
    {\LARGE{\textbf{Supplementary Materials}}}
\end{center}

\vspace{1em}

\begin{appendices}

\section{Summary of Content}
In \secref{sec:training-diagram}, we discuss and illustrate our adaptive framework as a diagram to show where the framework is applied during training (see \figref{fig:overview}). In \secref{sec:addressing_degenerate_alpha_gamma}, we discuss the possible issue of degenerate $\alpha_\tau$ and $\gamma$ and how our normalization scheme prevents it from occurring. In \secref{sec:sensitivity_hyperparameters}, we discuss our hyper-parameters and examine how sensitive they are to various values and whether they are able to generalize across datasets.
In \secref{sec:comparison-of-adaptive-frameworks}, we provide an extensive study to compare different adaptive frameworks (e.g. ``adaptive'' edge-awareness weights, \cite{wong2019bilateral}) and show that the proposed method performs the best. Then in \secref{sec:addressing-failure-modes}, we qualitatively show the failure modes in recent unsupervised depth completion methods and demonstrate how our framework addresses these issues (see \figref{fig:kitti-testing-results-ma-shivakumar}). Lastly, in \secref{sec:kitti-benchmark}, we show a comparison between unsupervised and supervised methods and how our approach can help close the gap between the two learning paradigms. We additionally show screen captures of the unsupervised KITTI benchmark at the time of submission.

\begin{figure}[t]
    \centering
    \includegraphics[width=1.00\linewidth]{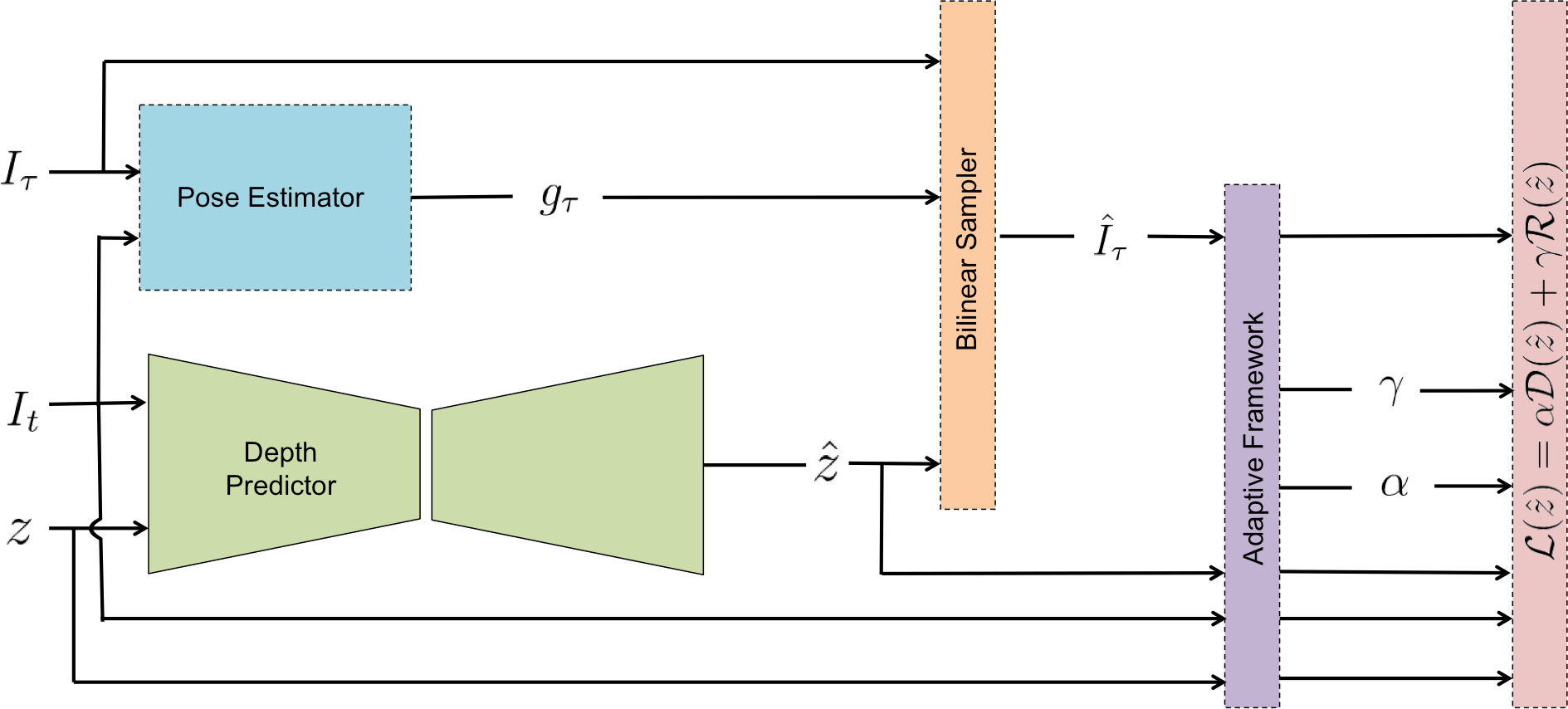}
    \caption{\textit{Diagram of the training pipeline using our framework.} Given the predicted depth $\hat{z}$, sparse depth $z$, image $I_{t}$ and its reconstructions $\hat{I}_{\tau}$, our framework (purple) is comprised of $\alpha$ (consists of $\alpha_{\tau}$ for $\tau \in \{-1, +1\}$) and $\gamma$ for adaptively weighting the data fidelity $\mathcal{D}$ and regularization $\mathcal{R}$ in the loss function (red). We note that our framework does not require any additional trainable parameters and incur no additional run-time complexity during inference; the only component required during inference is the depth predictor (green).}
    \label{fig:overview}
    \vspace{-1em}
\end{figure}

\section{Learning with Our Framework}
\label{sec:training-diagram}
In \figref{fig:overview}, we show where our adaptive framework is applied (purple block) during training. Given the predicted depth $\hat{z}$, sparse depth $z$, image $I_{t}$ and its reconstructions $\hat{I}_{\tau}$, we construct $\alpha$, which consists of $\alpha_{\tau}$ for $\tau \in \{-1, +1\}$, and $\gamma$ for adaptively weighting the data fidelity $\mathcal{D}$ and regularization $\mathcal{R}$ in the loss function. To optimize for this objective, we use alternating minimization where we first fix $\alpha$ and $\gamma$ and update $\hat{z}$, then subsequently, update $\alpha$ and $\gamma$ based on the residual given by the updated $\hat{z}$. Hence, a single training step consists of predicting $\hat{z}$, constructing $\alpha$ and $\gamma$ from residuals, computing the loss function, and backpropagation.

\section{Addressing degenerate $\alpha_\tau, \gamma$}
\label{sec:addressing_degenerate_alpha_gamma}
Because the data fidelity and regularization terms are controlled by $\alpha_\tau$ and $\gamma$, it would be problematic if they arrive at degenerate solutions e.g. all-zero estimates. However, our adaptive framework is designed to avoid such degeneracy issue. This is the intent of our residual normalization scheme (Eqn. 5, main text). To demonstrate this, let's consider $\alpha_\tau$ (Eqn. 6, main text):
\begin{equation}
    \alpha_{\tau}(x) = 1 - \frac{1}{1 + \exp(-(a\rho_{\tau}(x)-b))}, \nonumber
\end{equation}
for $\alpha_\tau$ to be all zeros, $\rho_\tau$ need to be very large, which can happen if $\sigma_\tau^2$ (Eqn. 4, main text) is very small. Yet, if $\sigma_\tau^2$ is small, then all of the residuals $\delta_\tau(x)$ (Eqn. 3, main text) would have similar values and therefore will be close to the mean $\mu_\tau$. Hence, the centered residuals, i.e. $\delta_\tau(x) - \mu_\tau$, would be close to 0, counteracting the small variance.

Let's consider $\gamma_i$ and $\gamma_z$ (Eqn. 12 and 14, main text), 
\begin{equation}
    \gamma_i(x) = \exp(-c_i \, \mu_i \, \delta_i(x)) \text{ for } x \in \Omega, \nonumber
\end{equation}
\begin{equation}
    \gamma_z(x) = \exp(-c_z \, \mu_z \, \delta_z(x)) \text{ for } x \in \Omega_z, \nonumber
\end{equation}

Indeed, if residuals are very large, $\gamma_i$ and $\gamma_z$ will tend to zero, but this is intended by our framework and desired during early time steps. In the case where both $\gamma_i$ and $\gamma_z$ produce all zeros, then we will maximize data fidelity e.g. allow exploration to find correspondences and minimize reprojection loss. This will immediately cause $\gamma_i$ and $\gamma_z$ to become non-zero.

\begin{figure}[t]
    \centering
    \includegraphics[width=1.00\linewidth]{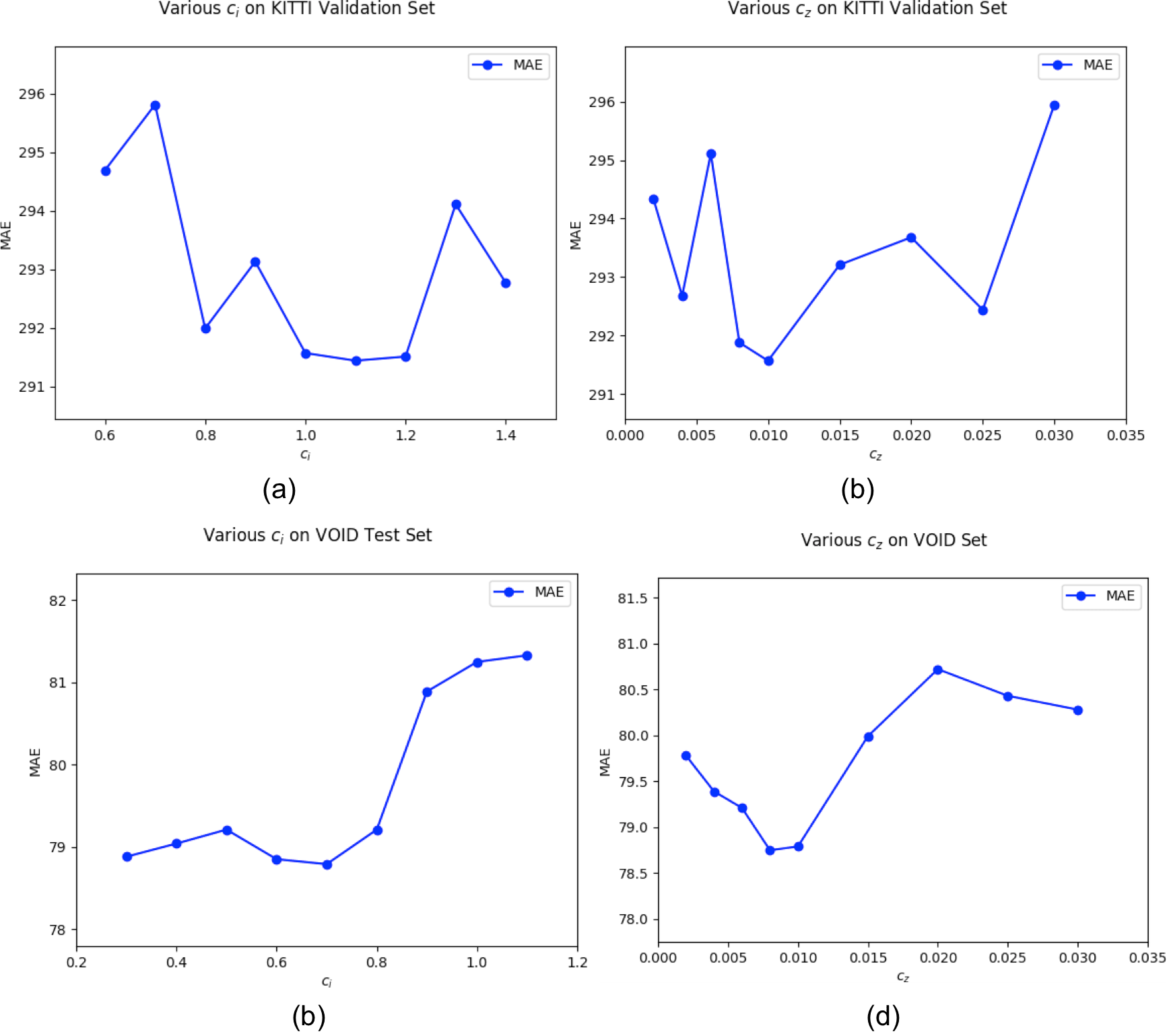}
    \caption{\textit{Sensitivity study on $c_i$ and $c_z$ on KITTI validation set and VOID test set.}. We apply our adaptive framework to VGG11 \cite{wong2020unsupervised} and train separate models for various $c_i$ and $c_z$ and evaluate the MAE metric. For KITTI, models are trained with \textbf{(a)} $c_i \in [0.60, 1.40]$ and $c_z = 0.01$, and \textbf{(b)} $c_i = 1.00$ and $c_z \in [0.002, 0.030]$. For VOID, models are trained with \textbf{(c)} $c_i \in [0.30, 1.10]$ and $c_z = 0.01$, and \textbf{(d)} $c_i = 0.70$ and $c_z \in [0.002, 0.030]$. Our framework is not too sensitive to $c_i$ and $c_z$. For KITTI, maximum change in MAE is $\approx$1.5\% and for VOID, it is $\approx$3.1\%.}
\label{fig:sensitivity_studies_ci_cz_kitti_void}
\end{figure}

\section{On sensitivity of hyper-parameters}
\label{sec:sensitivity_hyperparameters}
While our method adaptively weights data fidelity with $\alpha_\tau$ and regularization with $\gamma$ without incurring additional trainable parameters, we do introduce several hyper-parameters -- namely, $a_0$, $b_0$, $\epsilon$, $c_i$ and $c_z$. However, because these hyper-parameters are dependent on image intensity and depth values, most of them are fixed across different datasets. For instance, because image intensity range is normalized between $[0, 1]$, the same $a_0$, $b_0$ and $\epsilon$ are used across all datasets e.g. VOID, KITTI and, in \tabref{tab:nyuv2_test_set} NYUv2. The only hyper-parameters that may need adjustments are $c_i$ and $c_z$. 

Because of the ratios of max to min values (dynamic range) for KITTI (1m to 100m) and VOID (0.1m to 10m) are both coincidentally $100\times$, we choose $c_z = 0.01$ for both. Yet, the dynamic range is usually similar for datasets within a domain and so it does not need to be tuned for each dataset (see \tabref{tab:nyuv2_test_set}). Finally, $c_i$ was decreased from 1.0 for KITTI to 0.70 for VOID to allow for more regularization (less aggressive discounting) due to textureless surfaces. This is only needed to get better performance. So, not having optimal hyper-paramters will not ``break'' the framework nor hinder unsupervised/self-supervised learning. 

Nonetheless, in \figref{fig:sensitivity_studies_ci_cz_kitti_void}, we show sensitivity studies on $c_i$ and $c_z$ for VGG11 \cite{wong2020unsupervised} to demonstrate that our framework is not too sensitivity to these two hyper-parameters. For KITTI, maximum change in MAE is $\approx$1.5\% (from 291.57 to 295.95) and for VOID, it is $\approx$3.1\% (from 78.79 to 81.33). If we were to use KITTI hyper-parameters ($c_i = 1.00$ and $c_z = 0.01$) for VOID, we will obtain an MAE of 81.25. So, despite not using the best hyper-parameters, we still improve \cite{wong2020unsupervised}.

\begin{table}[t]
    \caption{Generalizing hyper-parameters from VOID to NYUv2}
    \centering
    \setlength\tabcolsep{3.5pt}
    \begin{tabular}{l c c c c c c}
        \midrule 
        Method & Dataset & MAE & RMSE & iMAE & iRMSE \\ 
        \midrule
        VGG11 \cite{wong2020unsupervised} & 
        NYUv2 & 131.71 & 223.34 & 30.04 & 52.67 \\
        \midrule
        VGG11 \cite{wong2020unsupervised} + $\alpha_\tau, \gamma$ & 
        NYUv2 & 122.86 & 202.52 & 26.89 & 46.71 \\
        \midrule
        VGG11 \cite{wong2020unsupervised} + $\alpha_\tau, \gamma$ & 
        VOID, NYUv2 & 120.11 & 200.08 & 25.40 & 46.13 \\
        \midrule
    \end{tabular}
    \begin{tablenotes}
        \textit{Training VGG11 on NYUv2 with fixed hyper-parameters from VOID.} Row 1: VGG11 trained on NYUv2 without our framework as a baseline. Row 2: VGG11 trained with our framework using $c_i=0.70$ and $c_z=0.01$. Row 3: VGG11 pretrained on VOID using our framework using $c_i=0.70$ and $c_z=0.01$ and trained on NYUv2 with the same hyper-parameters. The hyper-parameters generalizes well from VOID to NYUv2 and yields consistent improvements. Pretraining on VOID and then training on NYUv2 also does not require changing hyper-parameters.
    \end{tablenotes}
\label{tab:nyuv2_test_set}
\end{table}

Because our hyper-parameters are not dataset specific, we can apply the same hyper-parameters to different datasets within a domain e.g. indoor, outdoor. For this we will consider the indoor datasets: VOID \cite{wong2020unsupervised} and NYUv2 \cite{silberman2012indoor}. To demonstrate that our hyper-parameters are not dataset specific, we will use hyper-parameters set for VOID to train a model on NYUv2. We show quantitative results in \tabref{tab:nyuv2_test_set} where we train VGG11 \cite{wong2020unsupervised} on NYUv2, with (using VOID hyper-parameters, $c_i = 0.70$ and $c_z = 0.01$) and without our framework; we also train a VGG11 model, that was pretrained on VOID with our framework, without changing the hyper-parameters on NYUv2. As we can see, the hyper-parameters generalize well from VOID to NYUv2 and yield consistent improvements. Pretraining on VOID first and then training on NYUv2 also does not require us to change the hyper-parameters. Hence, our method is not limited by the choice of hyper-parameters -- tuning them will bring even more improvements.

\begin{figure*}[!h]
    \centering
    \includegraphics[width=0.95\linewidth]{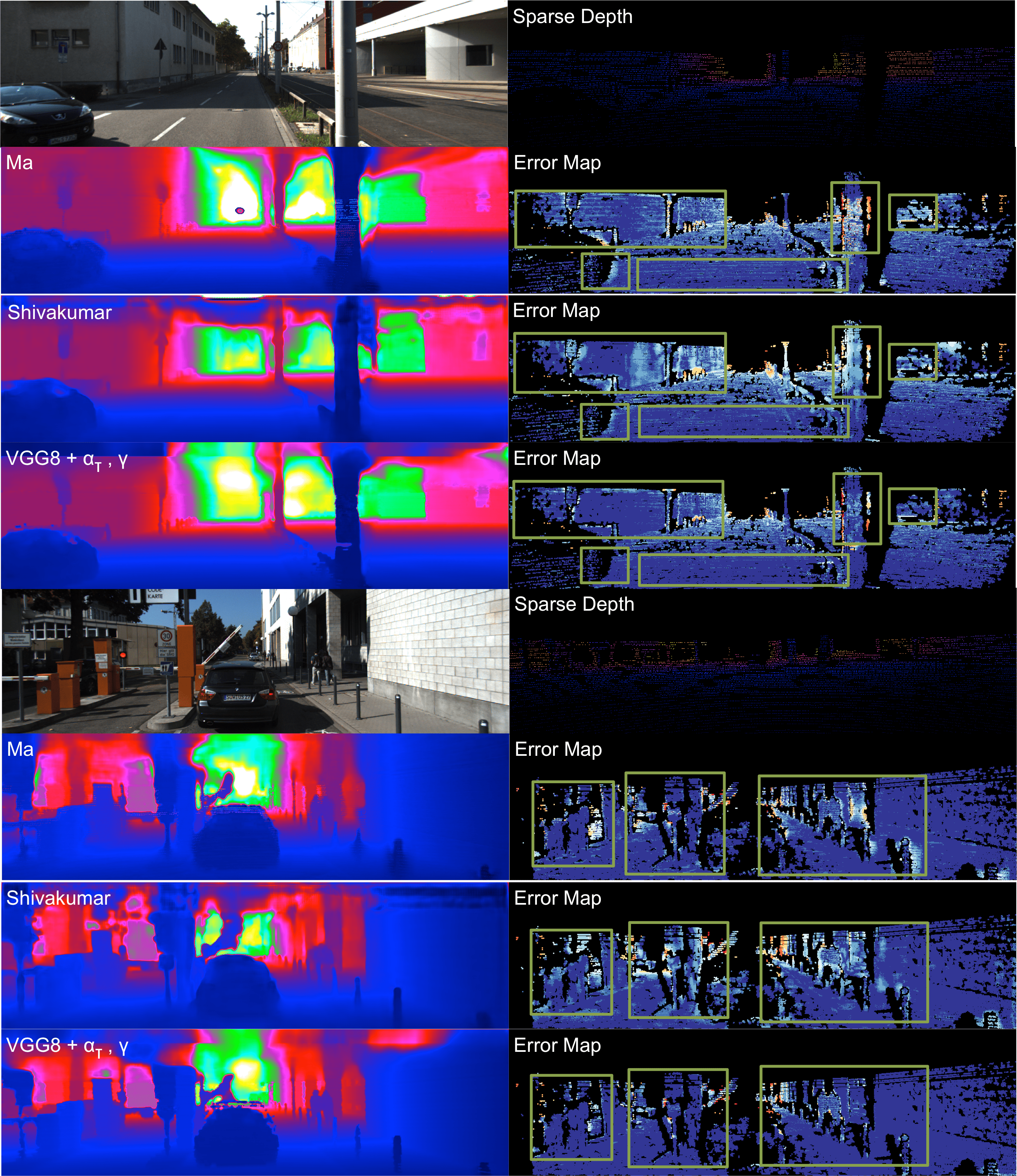}
    \caption{\textit{Head-to-head comparisons with \cite{ma2019self,shivakumar2019dfusenet}.} Like \cite{wong2020unsupervised}, \cite{ma2019self,shivakumar2019dfusenet} minimize a similar loss function which produces solutions with failure modes in low-textured regions (generally correspond to walls and roads, mark in green) and objects boundaries. Our approach specifically addresses those issues using our adaptive weights, $\alpha_\tau$ and $\gamma$. $\alpha_\tau$ addresses occlusions and dis-occlusions and $\gamma$ modulates regularization (in this case, local smoothness and forward-backward/left-right consistencies) to allow the model to find correct correspondences before imposing regularization. Our model is consistently better (more blue regions). Regions for detail comparisons are highlighted in green in the error map.}
    \label{fig:kitti-testing-results-ma-shivakumar}
\end{figure*}

\begin{table}
    \centering
    \caption{Comparison of adaptive frameworks on VOID test set}
    \setlength\tabcolsep{6pt}
    \begin{tabular}{l c c c c}
        \midrule
        Method & MAE & RMSE & iMAE & iRMSE \\ \midrule
        Ma \cite{ma2019self}
        & 178.85 & 243.84 & 80.12 & 107.69 \\ \midrule
        Ma \cite{ma2019self} + \cite{wong2019bilateral}
        & 178.85 & 243.84 & 80.12 & 107.69 \\ \midrule
        Ma \cite{ma2019self} + $\gamma$
        & 160.38 & 229.62 & 68.03 & 95.73 \\ \midrule
        Ma \cite{ma2019self} + $\alpha_\tau, \gamma$, w/o $\nabla I$
        & \textbf{152.49} & \textbf{221.11} & 67.13 & \textbf{92.80} \\ \midrule
        Ma \cite{ma2019self} + $\alpha_\tau, \gamma$, w/ $\nabla I$
        & 157.24 & 223.37 & \textbf{66.89} & 93.75 \\ \midrule
        VGG11 \cite{wong2020unsupervised}  
        & 85.05 & 169.79 & 48.92 & 104.02 \\ \midrule
        VGG11 \cite{wong2020unsupervised} + \cite{wong2019bilateral}
        & 81.98 & 152.53 & 45.99 & 96.31  \\ \midrule
        VGG11 \cite{wong2020unsupervised} + $\gamma$
        & 78.20 & 140.86 & 45.41 & 85.20 \\ \midrule
        VGG11 \cite{wong2020unsupervised} + $\alpha_\tau, \gamma$, w/o $\nabla I$
        & \textbf{76.29} & 136.61 & \textbf{42.17} & 81.07 \\ \midrule
        VGG11 \cite{wong2020unsupervised} + $\alpha_\tau, \gamma$  w/ $\nabla I$
        & 78.79 & \textbf{135.93} & 43.62 & \textbf{78.22} \\
        \midrule
    \end{tabular}
    \begin{tablenotes}
        \textit{Quantitative results of different adaptive methods.} We apply single image based \cite{wong2019bilateral} to \cite{ma2019self} and VGG11 \cite{wong2020unsupervised}. Our method consistently outperforms \cite{wong2019bilateral} because \cite{wong2019bilateral} only considers a single frame and does not consider depth as an adaptive signal (rows 1, 7). In fact, \cite{wong2019bilateral} performs worse than just using $\gamma$ because $\gamma$ leverages inter-frame residuals as guidance. We also study the case where we remove ``adaptive'' image gradient weights (i.e. ``edge-awareness'' term, $\nabla I$) commonly used in local smoothness terms (rows 4, 9). Compared to rows 5, 10, results are similar, but our framework actually improves slightly on some metrics when not considering the ``edge-awareness'' term. We believe this is because image gradients are static with respect to the image and noisy when trying to capture object boundaries; whereas, $\gamma$ captures object boundaries based on reconstruction residual and gradually regularized via an annealing process.
    \end{tablenotes}
    \label{tab:comparison-adaptive-framework}
\end{table}

\begin{figure*}[!ht]
    \centering
    \includegraphics[width=0.95\linewidth]{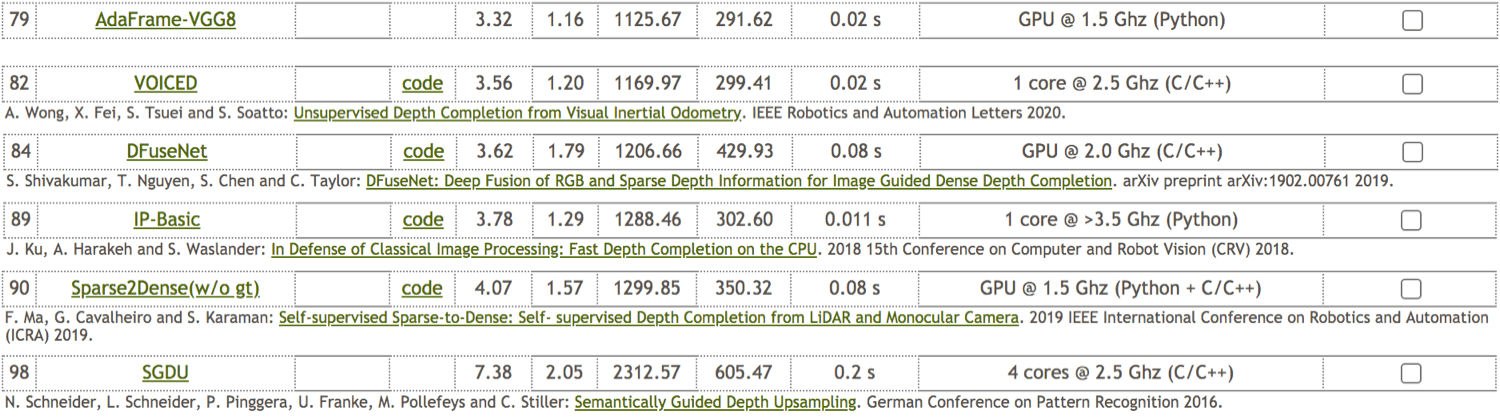}
    \caption{\textit{Screenshot of the KITTI depth completion benchmark.} \cite{yang2019dense} did not release their unsupervised results on the benchmark, but reports it in their paper. Our approach (AdaFrame-VGG08) achieves the state of the art on the KITTI unsupervised depth completion benchmark. Note: VOICED refers to \cite{wong2020unsupervised} using their VGG11 architecture. Despite applying our framework to VGG8, we still outperform VGG11.}
    \label{fig:kitti-benchmark-screenshot}
    \vspace{-1em}
\end{figure*}

\section{Comparison of adaptive frameworks}
\label{sec:comparison-of-adaptive-frameworks}
Because we are the first adaptive framework for the depth completion task, there are no methods for direct comparison. Hence, we have chosen an adaptive method \cite{wong2019bilateral} from the single image depth prediction literature. We provide a comparative study on VOID by applying the weighting scheme of \cite{wong2019bilateral} to \cite{ma2019self} and VGG11 \cite{wong2020unsupervised} in \tabref{tab:comparison-adaptive-framework}. Overall, \cite{ma2019self} with \cite{wong2019bilateral} performs worse than using \cite{ma2019self} + our $\gamma$ alone. We believe that this is due to \cite{wong2019bilateral} not accounting for sparse depth residual when determining the degree of regularization.

Additionally, we also consider the commonly used ``adaptive'' weighting scheme based on image gradients -- ``edge-awareness'' weights that are usually applied to the local smoothness term. We show the effect of our framework on \cite{ma2019self} and VGG11 \cite{wong2020unsupervised} \textit{with} the edge-awareness term (rows with ``w/ $\nabla I$'') and \textit{without} it (rows with ``w/o $\nabla I$''). The goal of the edge-awareness term is to reduce the amount of regularization across object boundaries in order to counteract oversmoothing. As seen in rows 4, 5 and 9, 10 in \tabref{tab:comparison-adaptive-framework}, the results are similar, but in the case of MAE, removing $\nabla I$ seems to improve performance. This may be due to the noisiness of $\nabla I$ (since it is directly utilizing the image gradients which are naturally noisy), which leads to undesirable regularization. We note that $\gamma$ naturally reduces regularization along object boundaries and provides a similar effect as $\nabla I$ with less noise towards convergence (as seen in Fig. III of the main text).

\begin{table}[t]
    \centering
    \setlength\tabcolsep{9pt}
    \caption{Supervised KITTI depth completion benchmark}
    \begin{tabular}{l c c c c}
        \midrule 
        Method & MAE & RMSE & iMAE & iRMSE \\ \midrule
        Chodosh \cite{chodosh2018deep}
        & 439.48 & 1325.37 & 3.19 & 59.39 \\ \midrule
        Dimitrievski \cite{dimitrievski2018learning}
        & 310.49 & 1045.45 & 1.57 & 3.84 \\ \midrule 
        VGG8 + $\alpha_{\tau}, \gamma$
        & \textit{291.62} & \textit{1125.67} & \textit{1.16} & \textit{3.32} \\
        \midrule
        Ma \cite{ma2019self} 
        & 249.95 & 814.73 & 1.21 & 2.80	\\ \midrule
        Qui \cite{qiu2019deeplidar}
        & 226.50 & 758.38 & 1.15 & 2.56	 \\ \midrule
        Xu \cite{xu2019depth}
        & 235.73 & 785.57 & 1.07 & 2.52 \\ \midrule	
        Chen \cite{chen2019learning}   
        & 221.19 & 752.88 & 1.14 & 2.34 \\ \midrule
        Van Gansbeke \cite{van2019sparse}
        & 215.02 & 772.87 & 0.93 & 2.19	\\ \midrule
        Yang \cite{yang2019dense}
        & \textbf{203.96} & 832.94 & \textbf{0.85} & 2.10 \\ \midrule		
        Cheng \cite{cheng2019cspn++} 
        & 209.28 & \textbf{743.69} & 0.90 & \textbf{2.07} \\ \midrule
        \end{tabular}
    \begin{tablenotes}
        \textit{Comparing our method against \textbf{supervised} methods on the KITTI depth completion benchmark}. All results are taken from the online benchmark \cite{uhrig2017sparsity}. Methods are ordered based on all metrics rather than just RMSE (ordering of \figref{fig:kitti-benchmark-screenshot}). Note: \cite{ma2019self,yang2019dense} compete in both supervised and unsupervised benchmarks. We compare our \textbf{unsupervised} adaptive framework ($\alpha, \gamma$, italized) against \textbf{supervised} methods. VGG8 trained with our framework outperforms \cite{chodosh2018deep} across all metrics, \cite{dimitrievski2018learning} on MAE, iMAE, and iRMSE metrics and over \cite{ma2019self} on the iMAE metric.
    \end{tablenotes}
    \label{tab:supervised-kitti-benchmark-results}
\end{table}

\begin{table}
    \centering
    \caption{Unsupervised KITTI depth completion benchmark}
    \setlength\tabcolsep{10pt}
    \begin{tabular}{l c c c c}
        \midrule
        Method & MAE & RMSE & iMAE & iRMSE \\ \midrule
        Schneider \cite{schneider2016semantically}   
        & 605.47 & 2312.57 & 2.05 & 7.38 \\ \midrule
        Ma \cite{ma2019self}
        & 350.32 & 1299.85 & 1.57 & 4.07 \\ \midrule
        Ku \cite{ku2018defense} 
        & 302.60 & 1288.46 & 1.29 & 3.78 \\ \midrule
        Shivakumar \cite{shivakumar2019dfusenet}
        & 429.93 & 1206.66 & 1.79 & 3.62 \\ \midrule
        Yang \cite{yang2019dense}
        & 343.46 & 1263.19 & 1.32 & 3.58 \\ \midrule
        VGG8 \cite{wong2020unsupervised}  
        & \textit{304.57} & \textit{1164.58} & \textit{1.28} & \textit{3.66} \\ \midrule
        VGG11 \cite{wong2020unsupervised}  
        & 299.41 & 1169.97 & 1.20 & \textit{3.56} \\ \midrule
        VGG8 + $\alpha_{\tau}, \gamma$
        & \textbf{291.62} & \textbf{1125.67} & \textbf{1.16} & \textbf{3.32} \\
        \midrule
    \end{tabular}
    \begin{tablenotes}
        \textit{Quantitative results on the \textbf{unsupervised} KITTI depth completion benchmark}. Results are directly taken from the online benchmark \cite{uhrig2017sparsity}. VGG8 have been italicized for relative improvement comparison. Applying our adaptive framework to an under-performing VGG8 model results in a performance boost that allows it to outperform the state-of-the-art VGG11. Table II in main text shows that our framework can consistently improve unsupervised methods for depth completion.
    \end{tablenotes}
    \label{tab:unsupervised-kitti-benchmark-results}
\end{table}

\section{Addressing the Failure Modes of Static Uniform  Weighting with our Framework}
\label{sec:addressing-failure-modes}
In Fig. 5 of the main text, we compared qualitatively to VGG11 \cite{wong2020unsupervised}. As \cite{ma2019self,shivakumar2019dfusenet,wong2020unsupervised} minimize a similar objective function, they all have similar failure modes with regards to low textured regions (where regularization is needed) and object boundaries (occlusions and dis-occlusions). 

In \figref{fig:kitti-testing-results-ma-shivakumar}, we illustrate these failure modes by showing a head-to-head comparison between \cite{ma2019self,shivakumar2019dfusenet} and VGG8 trained with our framework. Yellow bounding boxes highlight regions of improvements in the images and depth maps and green bounding boxes correspond to those regions in the error map. We note that the boxes mark similar regions for both \cite{ma2019self} and \cite{shivakumar2019dfusenet}. This is due to  \cite{ma2019self,shivakumar2019dfusenet} minimizing a similar objective function (photometric consistency, sparse depth consistency, and local smoothness regularizer). We also note that VGG8 trained with our framework minimizes a similar loss function, except that the data fidelity and regularization terms are adaptively weighted by $\alpha_{\tau}$ and $\gamma$ (Sec. IV and V in main paper). As a result, we do not observe the same failure modes in \cite{ma2019self,shivakumar2019dfusenet} -- reducing error in low-textured regions and object boundaries and, thus, demonstrating the effectiveness of our approach. We note that these results are also consistent in Fig. 4 of the main text, where we show qualitative comparisons of VGG8 and VGG11 trained with our framework with the vanilla VGG8 and VGG11 \cite{wong2020unsupervised}. This is also apparent in Fig. 5 of the main text as well.

\section{KITTI Depth Completion Benchmark}
\label{sec:kitti-benchmark}
In \figref{fig:kitti-testing-results-ma-shivakumar}, we show qualitative comparisons between VGG8 trained with our framework, \cite{ma2019self} and \cite{shivakumar2019dfusenet}. In \figref{fig:kitti-benchmark-screenshot}, we show the ranking of our method on the unsupervised KITTI depth completion benchmark. \figref{fig:kitti-benchmark-screenshot} is constructed by taking individual screenshots of each of the competing methods on the unsupervised depth completion task and concatenating them together based on ranking. By applying our method (ScaffFusion) to a lower ranking method VGG8 \cite{wong2020unsupervised} (shown in \tabref{tab:unsupervised-kitti-benchmark-results}), we improved it beyond the state-of-the-art VGG11 \cite{wong2020unsupervised}.

In \tabref{tab:supervised-kitti-benchmark-results}, we show recent methods in the supervised KITTI depth completion task. We note that \cite{ma2019self,yang2019dense} compete in both supervised and unsupervised settings. In \tabref{tab:unsupervised-kitti-benchmark-results}, we show unsupervised methods on the KITTI benchmark. In both tables, the methods have been re-ranked based on all four metrics rather than just RMSE. We note that while supervised methods performs better in general, our approach closes the gap between supervised and unsupervised methods. VGG8 trained with our framework not only outperforms \cite{chodosh2018deep} across all metrics, \cite{dimitrievski2018learning} on MAE, iMAE and iRMSE, but also \cite{ma2019self} on iMAE and closing in on other metrics. These improvements are obtained without any additional trainable parameters, pre- or post-processing, nor any extra inference time. We hope that by applying our framework, unsupervised methods can eventually reach the performance of supervised methods by simply finding better local minima. 

While applying our approach to an underperforming VGG8 does improve its performance over VGG11 to achieve the state-of-the-art in unsupervised depth completion, we would like to highlight our improvement on the iRMSE metric. While all recent methods have wavered between 3.5 and 3.6, our approach significantly improves VGG8 in this category to 3.32. This metric measures outliers in regions close to the camera, which is important for navigation. We attribute this to our adaptive framework, which discounts occlusions and disocclusions while imposing regularization where necessary.

\end{appendices}

\end{document}